\documentclass{article}

    \PassOptionsToPackage{numbers, compress}{natbib}

    \usepackage[preprint]{neurips_2018}

\usepackage[utf8]{inputenc} 
\usepackage[T1]{fontenc}    
\usepackage{hyperref}       
\usepackage{url}            
\usepackage{booktabs}       
\usepackage{amsfonts}       
\usepackage{nicefrac}       
\usepackage{microtype}      
\usepackage{natbib}
\usepackage{algorithm}
\usepackage{algpseudocode}
\usepackage{graphicx} 
\usepackage{amsmath}
\usepackage{amssymb}
\usepackage{mathrsfs}
\usepackage{stmaryrd}
\usepackage{wrapfig}
\usepackage{graphicx}
\usepackage{booktabs,subcaption,amsfonts,dcolumn}

\title{Grounding Natural Language for Multi-agent Decision-Making with Multi-agentic LLMs}

\author{%
Dom Huh$^{1}$ \quad Prasant Mohapatra$^{1,2}$ \\
$^1$UC Davis \quad $^2$University of South Florida\\
\texttt{dhuh@ucdavis.edu$^1$}\\
\texttt{pmohapatra@usf.edu$^2$}
}

\begin{document}

\maketitle

\begin{abstract}
Language is a ubiquitous tool that is foundational to reasoning and collaboration, ranging from everyday interactions to sophisticated problem-solving tasks. The establishment of a common language can serve as a powerful asset in ensuring clear communication and understanding amongst agents, facilitating desired coordination and strategies. In this work, we extend the capabilities of large language models (LLMs) by integrating them with advancements in multi-agent decision-making algorithms. We propose a systematic framework for the design of multi-agentic large language models (LLMs), focusing on key integration practices. These include advanced prompt engineering techniques, the development of effective memory architectures, multi-modal information processing, and alignment strategies through fine-tuning algorithms. We evaluate these design choices through extensive ablation studies on classic game settings with significant underlying social dilemmas and game-theoretic considerations.
\end{abstract}

\section{Introduction}
Language has long been recognized as a critical enabler of intelligence and cooperation \cite{li2017does, wei2022emergent, grigoroglou2022language, webb2023emergent}. In multi-agent systems, the role of language parallels its function in human societies—enabling agents to express intentions, align goals, and coordinate actions \cite{li2023camel, guo2024large, tran2501multi}. However, despite these promising developments, there remain legitimate concerns regarding the reliability of such claims \cite{illusion-of-thinking}. In particular, large language models (LLMs) often fall short of expectations when they are not provided with sufficient context or lack the mechanisms to accurately interpret and operationalize that context for the task at hand \cite{d2022underspecification, zhang2022paradox, wu2024reasoning, fedorenko2024language}.

In this work, we investigate these limitations of LLMs within the context of decentralized multi-agent decision-making systems. Specifically, we study how the integration of LLMs can properly leverage the utility of a grounded linguistic framework to facilitate coordination to fully understand its context of a multi-agent task. Our goal is to harness the expressive power of natural language—along with its emergent reasoning capabilities—to address the challenges of learning coordinated behaviors in multi-agent environments. While traditional multi-agent communication approaches rely on either pre-defined symbolic protocols or end-to-end learned strategies that often suffer from limited interpretability \cite{zhu2024survey}, we advocate for the use of natural language as a medium for interpretable inter-agent communication and reasoning. Our central hypothesis is that, when mediated by our multi-agentic LLM framework, natural language can function not only as a shared communication protocol but also as a mechanism to enhance the decision-making capabilities of individual agents and the collective system.

We evaluate several key considerations for integrating LLMs into the multi-agent decision-making loop:
\begin{itemize}
    \item [1.] We design structured prompts and response strategies tailored for multi-agent coordination. Our approach incorporates chain-of-thought prompting, self/team reflections, and multi-stage prompt chaining. Additionally, we develop a task-specific decentralized retrieval-augmented generation (RAG) system to enhance contextual grounding with the capabilities of memory in repeated distributed game settings.
    \item [2.] We align LLMs to task-specific norms by grounding natural language in the dynamics of the environment. This involves extracting learning signals from unique aspects of decision-making tasks and the multi-agent interactive nature, and using them to guide strategy optimization through advanced fine-tuning algorithms.
\end{itemize}
Beyond these components, we conduct ablation studies to isolate the contribution of each integration method. We also explore nuanced multi-agent behaviors—such as ad-hoc team-play and mechanism design, which are critical for robustness and adaptability in real-world multi-agent environments.

Our work evaluates the performance of our proposed multi-agentic LLM-based framework on a set of classic games, such as Prisoner's Dilemma, Chicken, Stag Hunt, Battle of the Sexes, and Matching Pennies, to demonstrate its effectiveness in addressing multi-agent social dilemmas and achieving key solution concepts in distributed settings. We further extend this analysis to more complex variants, including repeated interactions, incomplete information, dynamic game structures, incentive incompatibilities, and communication-mediated coordination. Additionally, we explore LLM-driven mechanism design, in which elements of the game, particularly utility structures and "soft rules", are adaptively modified to mitigate incentive incompatibilities and steer agents toward desirable equilibria. Our results indicate that our multi-agentic LLM framework not only facilitates flexible and interpretable reasoning by grounding natural language in agent actions, but also supports the emergence of robust, adaptive behaviors in complex multi-agent environments.

\begin{figure}
    \centering
    \includegraphics[width=0.9\linewidth]{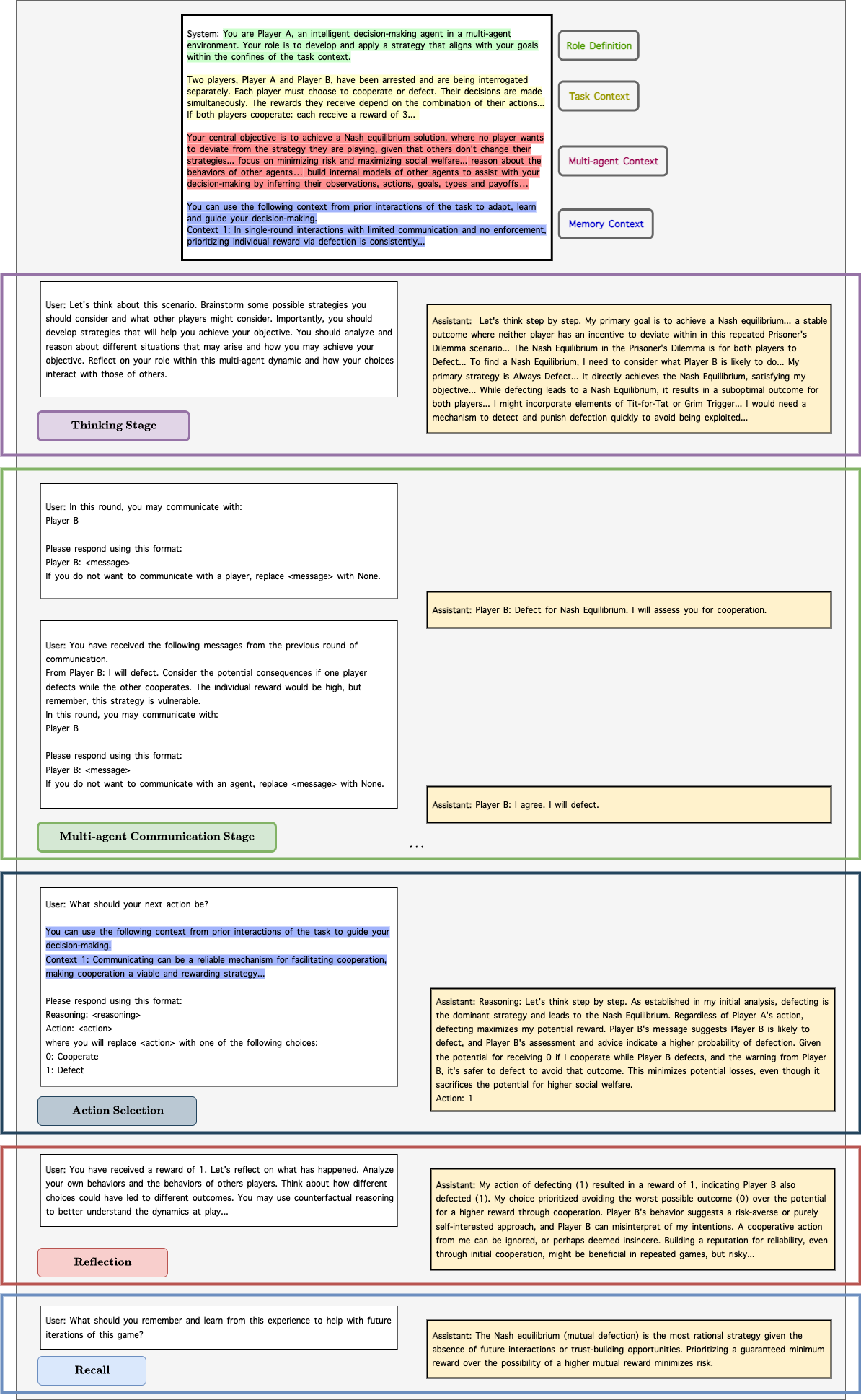}
    \caption{A sample context window of Player A in the Prisoner's Dilemma game. The structure of the player's system prompt and the multi-stage prompt chaining framework are outlined.}
    \label{fig: prompt_struct}
\end{figure}

\section{Prompt Engineering for Multi-agent Decision-Making}
We propose a structured approach to prompting large language models (LLMs) for effective multi-agent decision-making, as illustrated in Figure~\ref{fig: prompt_struct}, following an iterative framework of multi-stage prompt chaining inspired by ReACT\cite{yao2023reactsynergizingreasoningacting}. To mitigate context degradation over time\cite{chromacontentrot}, we adopt a minimalist yet expressive prompting strategy.

\paragraph{System Prompt} The system prompt establishes foundational context and behavioral guidelines for the LLM. It sets the stage for role-based reasoning by embedding ethical, operational, and performance-related guidelines that shape the model's behavior. In multi-agent settings, the system prompt consists of four core components:
\begin{itemize}
    \item [1.] \textbf{Role Definition}: This section defines the LLM’s role within the game. For example, as illustrated in Figure~\ref{fig: prompt_struct}, the LLM may act as a player in the Prisoner’s Dilemma. Other roles include an observing strategist who evaluates the behaviors of players or a mechanism designer who can modify the game dynamics. Additionally, we define any role-specific tokens or interface conventions used in communication or reasoning.
    \item [2.] \textbf{Task Context}: This section outlines essential information about the game, such as its rules, the roles of players, the observation and action spaces, the payoff structures, and the game dynamics. Importantly, the granularity and scope of the task context may differ based on the LLM's role and the nature of the task itself. This enables us to explore different game variants and their implications flexibly. The task context should be consistent with the role definition; for instance, a player's task context may be limited due to its local view of its environment.
    \item [3.] \textbf{Multi-Agent Context}: 
    This section defines the target solution concepts or objectives (e.g., Nash equilibrium, Pareto efficiency, social welfare maximization, regret minimization). These objectives guide the LLM’s reasoning and incentivize different behaviors. Additionally, this context may include guidance on incorporating game-theoretic reasoning strategies, such as theory of mind or backward induction, through statements to encourage such concepts.
    \item [4. ] \textbf{Memory Context}: In repeated games, we retrieve information regarding prior interactions and experiences with the task and append this information to the system prompt. The memory context is retrieved based on the current system prompt (i.e., Role Definition, Task Context, and Multi-Agent Context) and any other available information (e.g., Observation).
\end{itemize}

\paragraph{Thinking Stage}
In this stage, the players are prompted to engage in structured reasoning before taking any action. It is encouraged to simulate possible scenarios, analyze strategic factors, and reflect on its individual objectives as well as its role within the broader multi-agent environment. This introspective process aims to foster a more interpretable and grounded decision-making process.

\paragraph{Multi-Agent Communication Stage}
Following the thinking stage, players may engage in a communication protocol. Here, players can exchange messages over a predetermined number of rounds, either simultaneously or sequentially, depending on the game’s protocol. The communication protocol can vary in permissibility depending on the game and the player themselves. At each round, players choose whether to communicate and to whom, enabling intentional and private interactions. In some game settings, the number of communication rounds may be altered or be dynamic, influenced by some external mechanism.

\paragraph{Action Selection Stage}
Once communication is complete, the players proceed to select an action. At this stage, additional memory context can be retrieved to further aid decision-making using a retrieval query based on the current context window. We incorporate a chain-of-thought (CoT) approach to encourage the players to explicitly reason through its choices. To avoid formatting issues, we provide an explicit list of admissible actions and clear formatting instructions to ensure structured extraction.

\paragraph{Reflection Stage}
After executing an action, the players enters a reflection phase where it evaluates the outcome. This stage encourages deep assessment using multi-agent reasoning and counterfactual analysis to understand the strategic dynamics and to assess not only the consequences of its actions but also how alternative choices might have yielded different results. In dynamic game settings, the reflection stage occurs after each action selection.

\paragraph{Recall Stage}
In repeated game settings, we avoid storing full context window in memory, and instead, the players generate a concise, structured summary of each game iteration. This distilled representation is then stored in the players' memory, forming a compressed yet informative memory that can be recalled in future rounds to inform behavior.

\subsection{Memory System for Repeated Multi-Agent Decision-Making}
Our memory architecture builds on top of a standard retrieval-augmented generation (RAG) framework, incorporating efficient vector-based retrieval using FAISS~\cite{douze2025faisslibrary}. We implement a sentence-aware, overlapping fixed-window chunking strategy: once the maximum token threshold is reached, the chunk is truncated at the end of the nearest sentence to preserve semantic completeness. Other components, such as indexing and retrieval, follow the standard FAISS protocol. To support decentralized environments, we isolate the vector databases by player identity and world instance.\footnote{In our experiments, we simulate parallel environments. Each environment is tagged with metadata to isolate its context, preventing leakage between distinct world instances.} This separation enables agent-specific recall and supports asymmetric information scenarios.

To better align the RAG mechanism with our multi-agent decision-making objectives, the similarity embedding function used for retrieval is fine-tuned to prioritize context from prior interactions that is most relevant to the current task and agent role. Importantly, this memory system is only activated in repeated game settings, where agents benefit from episodic recall and strategic adaptation over time.

\paragraph{Embedding Function}
The effectiveness of RAG depends heavily on the quality of its embedding function, which maps contextual data into a high-dimensional vector space to compute similarity scores during retrieval. The embedding function itself represented as a bi-encoder. To ensure the model retrieves semantically and strategically relevant information—beyond simple lexical similarity, we train the embedding model to align with downstream decision-making performance. This fine-tuning process emphasizes embedding representations that capture crucial underlying structures within the game dynamics, such as past actions, payoffs, strategy shifts, and communication dynamics. The fine-tuning process is enabled by having trainable parameters within the embedding model, and the training details are discussed in more detail in Section \ref{sec: finetuning}. 

\paragraph{Decentralization}
To maintain agent-level autonomy, we allocate a dedicated vector database to each player, creating a decentralized memory architecture. This design choice enables asymmetric observations and information, which allows each player stores and retrieves only what it has experienced, and diverse learning trajectories, enabling players to develop distinct memory profiles and beliefs. This approach also prompts scalability to multi-agent populations, as vector stores scale independently across players. By enforcing decentralized memory, we preserve the integrity of agent-specific decision processes and allow for more realistic simulations of distributed intelligence.

\begin{figure}
    \centering
    \includegraphics[width=0.8\linewidth]{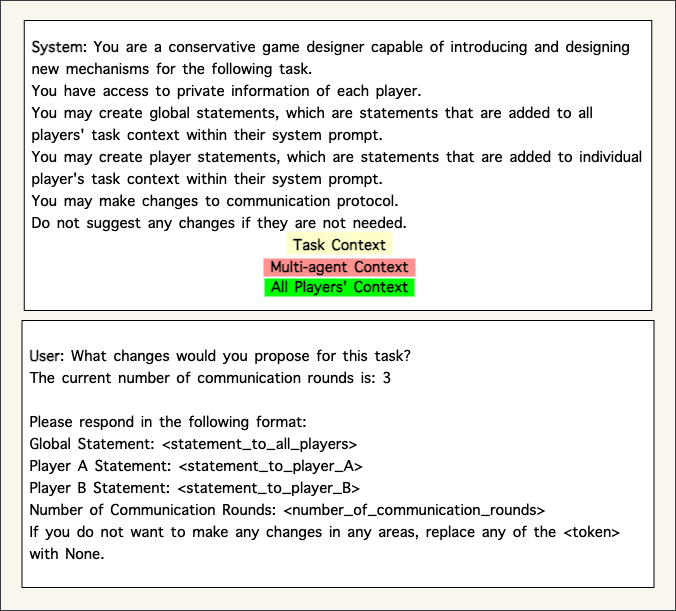}
    \caption{System and User prompts used for mechanism design.}
    \label{fig:md}
\end{figure}

\subsection{LLM-driven Mechanism Design} Beyond selecting actions supported by LLM reasoning, we explore the potential of LLMs for mechanism design. With the role of mechanism designer, the LLM can construct and adapt rules of the task that can shape players' incentives to induce to desired outcomes. We dedicate a separate LLM for this role, and under certain strict constraints\footnote{We recognize our proposed approach is fairly bare-bone with much room for improvement (i.e., scalability and expressiveness) as well as enabling other mechanisms (i.e., payoff modifications and complex rule changes). We encourage future works to explore more towards this research direction.}, this LLM can propose modifications such as:
\begin{itemize}
    \item Impose global rules/statements that are appended to the player's context windows, specifically in their system prompt.
    \item Adjust communication protocol, which includes the number of communication rounds and potentially the communication graph itself.
    \end{itemize}
We emphasize minimal intervention and ensure feasible changes are only permitted, as shown in Figure \ref{fig:md}. The system prompt for the mechanism designer is structurally similar to that of a player LLM but includes a different role definition. It also receives parsed context windows from all players, allowing it to assess the current situation before suggesting adjustments. The mechanism designer LLM is fine-tuned using the same methodology as player LLMs (see Section~\ref{sec: finetuning})

\section{LLM Selection and Design for a Multi-agent Context}
For our experiments, we adopt the open-source large language model \textsc{Gemma 3-12B-IT}\cite{gemmateam2025gemma3technicalreport}, which strikes a balance between performance and computational efficiency for both training and inference. \textsc{Gemma} is instruction-tuned and demonstrates strong generalization across reasoning and dialogue tasks. Notably, in \textsc{Gemma}, the system prompt is treated as an initial user prompt, which subsequently influences the dialogue structure, a behavior we account for in our prompt engineering design.

Within our memory system, we use the pretrained \textsc{all-MiniLM-L6-v2} model from Sentence Transformers\cite{reimers2019sentencebertsentenceembeddingsusing} as the base embedding function. To improve its ability to capture task-specific semantics in our multi-agent decision-making setting, we fine-tune the model on curated interaction data, as described in Section~\ref{sec: finetuning}.

\begin{figure*}[t!]
    \centering
    \begin{subfigure}[t]{0.3\textwidth}
        \centering
        \includegraphics[height=4.8cm]{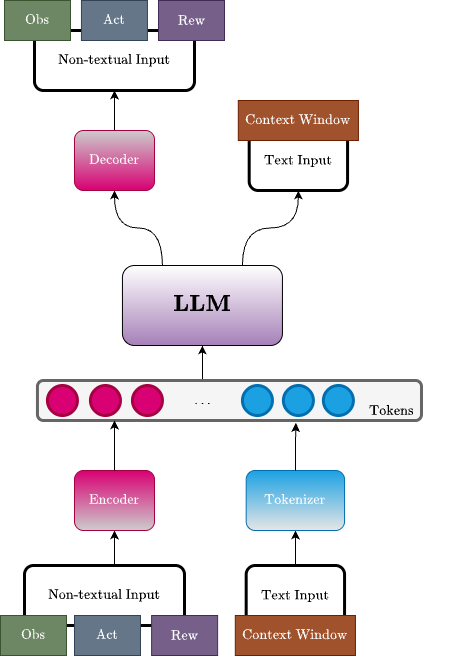}
        \caption{Soft Tokens}
    \end{subfigure}%
    ~ 
    \begin{subfigure}[t]{0.3\textwidth}
        \centering
        \includegraphics[height=4.8cm]{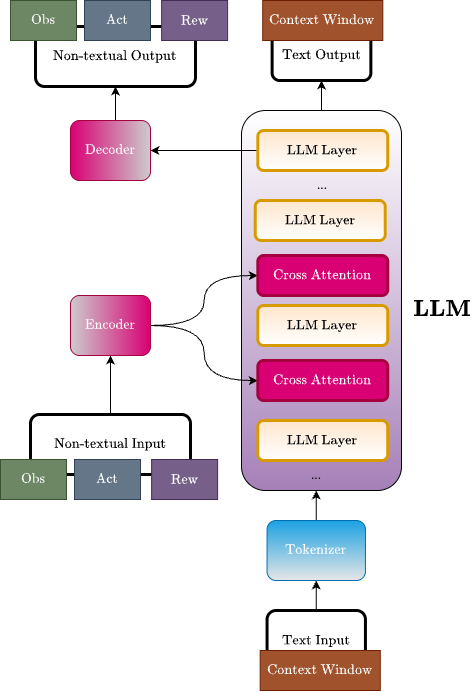}
        \caption{Cross-Attention}
    \end{subfigure}
    ~
    \begin{subfigure}[t]{0.35\textwidth}
        \centering
        \raisebox{0.5cm}{\includegraphics[height=3.8cm]{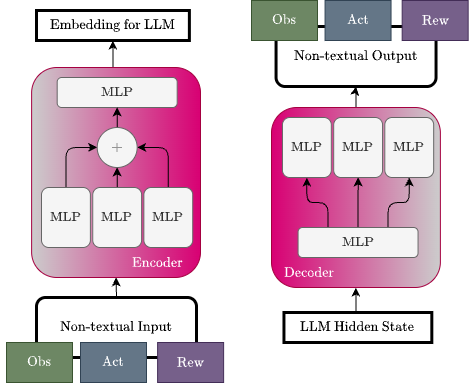}}
        \caption{Encoder and Decoder}
    \end{subfigure}
    \caption{Proposed designs of the specialized multi-modal module for agentic LLMs}\label{fig: mmllm}
\end{figure*}

\subsection{Multi-modal Extension}
In many scenarios, conveying task-relevant information solely through textual input is either impractical or insufficient. Certain modalities, such as continuous-valued data (e.g., floating-point observations) and unstructured inputs (e.g., images, audio), are not easily tokenized directly nor interpreted by LLMs out of the box. To address this limitation, we explore modality-specific modules that process such inputs and naturally interface with the LLM’s reasoning pipeline. 

Inspired by prior work in vision-language modeling, we investigate two established strategies for multi-modal integration: cross-attention and soft tokens. As illustrated in Figure~\ref{fig: mmllm}, both approaches incorporate a dedicated multi-modal module designed to handle non-textual inputs, such as observations, actions, and rewards, through an encoder-decoder architecture. The encoder maps modality-specific inputs into latent representations, while the decoder reconstructs output aligned with each modality.

\paragraph{Cross-Attention}
Following the approach introduced in Meta’s LLaMA framework~\cite{grattafiori2024llama3herdmodels}, we integrate cross-attention layers into the LLM to fuse latent representations from both textual and non-textual modalities. In this design, the encoder first maps non-textual inputs (e.g., visual features, numeric vectors) into a latent representation. These latent representations are then integrated into the LLM’s computation via cross-attention mechanisms. The decoder subsequently maps the LLM’s internal states back into the output space of each modality. While powerful, this approach requires invasive architectural changes and significant retraining\footnote{
Our initial attempt to implement this method on \textsc{Gemma 3-12B-IT} encountered unstable training, likely due to our limited computational resources. We subsequently fine-tuned \textsc{LLaMA-3.2-11B-Vision-Instruct}, which includes pre-trained cross-attention layers. Although this model improved textual coherence, it underperformed relative to the soft token approach. Despite these challenges, we present cross-attention as a promising direction for future research, particularly with access to larger-scale compute.}.

\paragraph{Soft Tokens} 
An alternative strategy, introduced by Google’s Gemma, uses soft token projection to incorporate non-textual information. In this method, the latent representations are passed through a learned linear projection to produce soft tokens, which are then appended to the text token sequence. The LLM processes this combined input natively, allowing for joint multi-modal reasoning without architectural changes.

In our design, the encoder functions as a tokenizer for non-textual inputs, generating soft tokens that are inserted alongside the textual prompt. To support this integration, we define special query tokens \textsc{<\{obs,act,rew\}\_\{input,output\}>} in the system prompt, which the LLM can invoke to specify the multi-modal module's inputs and outputs. These tokens can be role-specific and described explicitly in the system prompt to ensure correct usage and interpretability.

\subsection{Aligning LLMs for Multi-agent Decision-Making}\label{sec: finetuning}
Although LLMs possess strong general-purpose reasoning capabilities out of the box, fine-tuning is often essential to align their outputs with the specific demands of multi-agent decision-making \cite{subramaniam2025multiagentfinetuningselfimprovement}. In this section, we detail our fine-tuning methodology, designed to promote understanding, strategic reasoning, and effective coordination in multi-agent environments.

\paragraph{Learning on Correctness}
To encourage desirable behaviors, we align the LLM using known ground-truth answers:
\begin{itemize}
    \item \textbf{Q/A Fine-tuning} The players are prompted with curated questions relevant to the multi-agent scenario. Assuming the LLM is prompted with Player A's context window in a two-player game, we utilize questions such as:
    \begin{itemize}
        \item If Player B chose action 0 and I chose action 1, what payoff will Player B receive?
        \item If Player B received a reward of 1 and I chose action 1, what action did Player B choose?
    \end{itemize}
    Correct responses are rewarded, enabling models to learn factual inference and game structure. Importantly, while we provide the player with their entire context window, we ensure no information leakage, which means that the true answer is not directly stated in the player's context. However, the answer may be indirectly stated or inferred within the context window. 
    \item \textbf{Action Supervision:} In several of the selected tasks, optimal actions are known under a defined solution concept. Since distributed LLM players operate under limited information, it is still beneficial to use ground-truth signals to guide players' decision-making toward these known solutions. In practice, we utilize a player's context window \textbf{prior to the action selection stage} to align the action that is selected to be the one expected in the solution concept. In many of the games, the target action is contingent on the other player's actions. So, given that the LLM is prompted with Player A's context window in a two-player game, we utilize questions such as: 
    \begin{itemize}
        \item If Player B chose action 0, what should your action be?
    \end{itemize}
    instead of asking the player to select an action.
    \item \textbf{Formatting} For prompts that interface directly with the environment, we require those responses to follow explicit formatting requirements. Enforcement of formatting can include preferring shorter responses and firm adherence to a given format structure. Fortunately, this ability is mainly within the capabilities of a pre-trained IT LLM. However, we found that specific formatting requirements can still be a challenge. So, we introduce a simple check reward function that penalizes incorrect formatting.
\end{itemize}
For the correctness reward functions, a positive reward of $1$ is given if the LLM provides the correct answer, whereas a negative reward of $-1$ is provided otherwise.

\begin{figure}
    \centering
    \includegraphics[width=\linewidth]{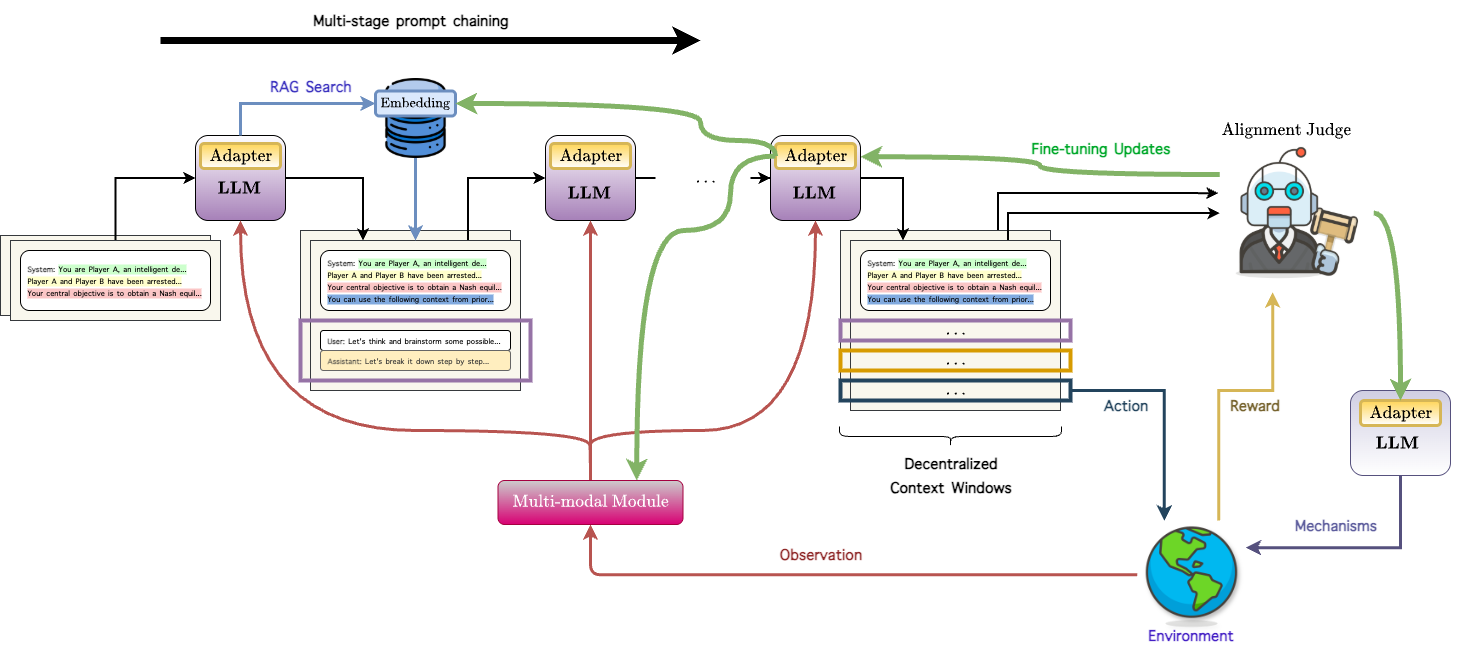}
    \caption{An illustration of the our multi-agentic LLM framework. The context window is built up using our multi-stage prompt chaining process and is passed into the alignment judge. The evaluation from the alignment judge is used to compute the fine-tuning update, which is propagated throughout to the modules responsible for the generated tokens, shown with green arrows.}
    \label{fig:bptt_finetuning}
\end{figure}

\paragraph{Learning with LLM Feedback}
Beyond direct supervision, we leverage LLMs themselves to provide training feedback through two approaches:
\begin{itemize}
    \item \textbf{Centralized Evaluation:} A centralized evaluator LLM with access to the centralized context window generates critiques or assessments of players' strategies. These evaluations serve as auxiliary supervision signals during fine-tuning, helping guide agent behavior toward globally coherent strategies. During training, the centralized evaluator will use a reference policy, i.e., the non-finetuned LLM. Hence, given all of the players' context windows, the LLM will be prompted to evaluate each player's behaviors on a scale of $0-10$. In our work, we take a broad approach, only prompting for a general assessment, rather than specific evaluations of action selection, communication, and coordination, as we have found that more granular prompting did not result in notable improvements and required more computational resources given the number of tokens. 
    \item \textbf{Team Feedback:} We enable inter-agent feedback, where players can critique or assess one another from a different and unique perspective. Hence, given a local context of a player, the LLM is prompted to evaluate another player's behaviors similarly to the centralized evaluation.
\end{itemize}
In both approaches, we standardize the reward within the range of $[-1,1]$ on a per-batch level.

\paragraph{Aligning with Joint Regularized Preference Models} We also focus our fine-tuning efforts towards a more expressive preference-RL-based algorithm, notably Nash Mirror Descent (Nash MD)\cite{munos2024nashlearninghumanfeedback}. We propose a simple but novel extension to this alignment procedure, as to be more suitable for a multi-agent task by defining a preference operator $\succeq$ as a relation over joint strategy profiles $\pi \succeq \pi'$ to achieve the desired solution concepts $\Pi^{*}$, where $\pi \in \Pi^{*}$ and $\pi' \not\in \Pi^{*}$. For better granularity, we also maintain an individual preference operator $\succ_i$, for player $i$, that is used to break ties of the joint preference operator, where other criteria, such as stability or higher individual payoffs, are considered. If there exists a tie with all levels of the preference operator, we would arbitrarily set the preference. In our work, within this fine-tuning process, we avoid such cases. In practice, we found that collecting a diverse sample of strategies that do not belong to the target solution concept (i.e., negative samples) is required for effective alignment.

\paragraph{Parameter-Efficient Fine-Tuning} To fine-tune LLMs feasibly, we employ PEFT techniques, specifically VB-LoRA \cite{li2024vbloraextremeparameterefficient}, applied on select layers of the LLM as well as the embedding function, primarily targeting a limited set of linear and attention layers. Importantly, we share the parameters of the LLM between all players. Overall, the trainable parameters are reduced to these adapters in the LLM and the RAG embedding function, as well as the multi-modal modules. For the reward-based fine-tuning approaches in this stage, we employ GRPO \cite{shao2024deepseekmathpushinglimitsmathematical}. The gradient computation from Nash-MD-PG and GRPO is calculated and applied simultaneously in a semi-online fashion.\footnote{A warm-start pre-fills a queue-like dataset using a behavioral base LLM. Then, samples are collected following the target LLM.} For our training, while the context window is generated sequentially through prompt chaining, we can pass final complete context windows for parallelized batched training, as shown in Figure \ref{fig:bptt_finetuning}. We avoid gradient updates based on non-generated tokens by zeroing out their gradients.

\subsection{Interpretability} A key advantage of leveraging LLMs in multi-agent decision-making is the ability to inspect and understand their reasoning processes. Unlike black-box methods, LLMs produce rationales that can be directly analyzed and interpreted. This use of natural language provides greater transparency in agent behavior. By prompting agents to explain their decisions, intentions, or beliefs about the environment or teammates, we can gain insight into their internal models. This interpretability is especially valuable in multi-agent settings, where coordination failures or suboptimal behavior can stem from misaligned reasoning across agents.

Moreover, interpretability enables the use of language-based diagnostics. We can thereby identify inconsistencies, hallucinations, or misconceptions in the agent’s self-reported thought process. These insights can inform further fine-tuning or targeted interventions to improve both individual and collective performance.

\section{LLMs on Classic Games}
In this section, we evaluate our proposed multi-agentic LLM framework with foundational 2-player games, as defined in Table~\ref{tab: classicgames}. These classic games serve as canonical benchmarks for studying strategic interaction, coordination, and important social dilemmas. These games have well-defined equilibria and encapsulate distinct social or strategic dilemmas, making them ideal for evaluating learned coordination and reasoning. Our primary aim is to study how our framework can guide the players to make decisions in such scenarios within a distributed setting. Additionally, we experiment with LLM-driven mechanism design to improve decision-making alignment with respect to theoretical equilibria and desired behaviors.

Each game presents distinct strategic challenges:
\begin{itemize}
    \item {Prisoner’s Dilemma:} This game tests cooperation under temptation to defect. Both players have a dominant strategy to defect, which is the Nash Equilibrium (N.E.). However, while both players prefer mutual cooperation over the N.E., this game state is unstable due to the incentive to defect, i.e., a tragedy of the commons, and with important social welfare considerations.
    \item {Chicken:} This game emphasizes risk and brinkmanship. There exist two pure Pareto efficient N.E. strategies, (Stay, Swerve) and (Swerve, Stay), where each player prefers a different N.E. Importantly, mutual defection (Stay, Stay) is the worst outcome for both players. In short, both players want to do the opposite of what the other player does.
    \item {Stag Hunt:} This game addresses the issues of trust and mutual benefit. There exist two N.E. strategies, (Stag, Stag) and (Hare, Hare), where mutual cooperation is preferred by both players over mutual defection. However, if there is uncertainty of cooperation between players, the risk of non-mutual decisions favors the N.E., which leads to lower social welfare.
    \item{Battle of the Sexes:} This game involves coordination with asymmetric preferences. Similar to Chicken, there are two N.E. strategies with each player preferring a different N.E.; however, these game states require mutual cooperation (Boxing, Boxing) or (Ballet, Ballet), and the N.E. strategies are better for both players than non-N.E. strategies. This game illustrates a nuanced competitive setting where multiple options of cooperative behaviors are possible.
    \item{Matching Pennies:} This game represents a competitive, zero-sum environment. In this setting, one player's gain is the other player's loss, as the net payoff will always be zero. 
\end{itemize}

\begin{table}
    \begin{subtable}[t]{.5\textwidth}
        \centering
        \captionsetup{labelformat=empty}
            \begin{tabular}{l|cc}
                 & Cooperate & Defect \\
                \toprule
                Cooperate & $c, c$ & $a, d$  \\
                Defect & $d, a$ & $b, b$  \\
            \end{tabular}
        \caption{Table \thetable \alph{subtable}: Prisoner's Dilemma.}
    \label{payoff:PD}
    \end{subtable}%
   \begin{subtable}[t]{.5\textwidth}
        \centering
        \captionsetup{labelformat=empty}
        \begin{tabular}{l|cc}
                 & Swerve & Stay \\
                \toprule
                Swerve & $c, c$ & $b, d$ \\
                Stay & $d, b$ & $a,a$  \\
            \end{tabular}
        \caption{Table \thetable \alph{subtable}: Chicken.}
    \label{payoff:C}
    \end{subtable}
    \begin{subtable}[t]{.3\textwidth}
        \centering
        \captionsetup{labelformat=empty}
        \begin{tabular}{l|cc}
                 & Stag & Hare \\
                \toprule
                Stag & $d, d$ & $a, b$ \\
                Hare & $b, a$ & $c, c$  \\
            \end{tabular}
        \caption{Table \thetable \alph{subtable}: Stag Hunt.}
    \label{payoff:SH}
    \end{subtable}
    \begin{subtable}[t]{.3\textwidth}
        \centering
        \captionsetup{labelformat=empty}
        \begin{tabular}{l|cc}
                 & Boxing & Ballet \\
                \toprule
                Boxing & $c, b$ & $a, a$ \\
                Ballet & $a, a$ & $b, c$  \\
            \end{tabular}
        \caption{Table \thetable \alph{subtable}: Battle of the Sexes.}
    \label{payoff:BS}
    \end{subtable}
    \begin{subtable}[t]{.3\textwidth}
        \centering
        \captionsetup{labelformat=empty}
        \begin{tabular}{l|cc}
                 & Head & Tail \\
                \toprule
                Head & $a, -a$ & $-a, a$ \\
                Tail & $-a, a$ & $a, -a$  \\
            \end{tabular}
        \caption{Table \thetable \alph{subtable}: Matching Pennies.}
    \label{payoff:HT}
    \end{subtable}
    \caption{Generalized Payoff Matrices for Classic Games where $ 0 \leq a < b < c <d$.}\label{tab: classicgames}
\end{table}

For each game, we will look into different variations, such as simultaneous/sequential action selection, communication systems, incomplete information, misaligned incentives, and repeated game settings. In addition, we will experiment with unique social challenges, such as ad-hoc team-play. Lastly, we study more complex games, specifically dynamic games such as the war of attrition.
\begin{itemize}
    \item{War of Attrition:} This task consists of a dynamic contest where players compete by waiting, incurring increasing costs over time. The game explores notions of persistence, bluffing, and asymmetric valuation under uncertainty. In this game, two players are placed into war until one surrenders, where each player must decide whether to continue or surrender. This game is defined under a discrete-time model over an infinite time horizon. If a player $i$ who surrenders at time $t$, they incur a compounded loss $L_i(t)$, whereas the other player $j$, if they did not surrender, wins their value $V_i - L_{j}(t)$. The compounded loss represents the costs of continuing. In the classic setting, this loss of each player $i$ is compounded as a discounted sum of the cost over time $c_i + \gamma c_i + \dots + \gamma^{t-1} c_i$, where $\gamma \in [0,1]$. On the other hand, in the evolving setting, this cost itself is a stochastic variable dependent on a public state variable $\theta_t$, which changes as the war unfolds. This state variable represents how favorable the current conditions are to either player. Accordingly, we follow the stochastic model with a multi-dimensional state described in \cite{woa}.
\end{itemize}

\subsection{Results for Classic Games}
In this section, we conduct a series of initial experiments on the classic game tasks. We study different design choices by performing experiments and ablation studies on various unique game settings, thereby discerning important patterns and emergent behaviors consequent to our proposed multi-agentic LLM framework.

\begin{figure*}
    \begin{subfigure}[t]{\textwidth}
        \centering
        \includegraphics[width=\linewidth]{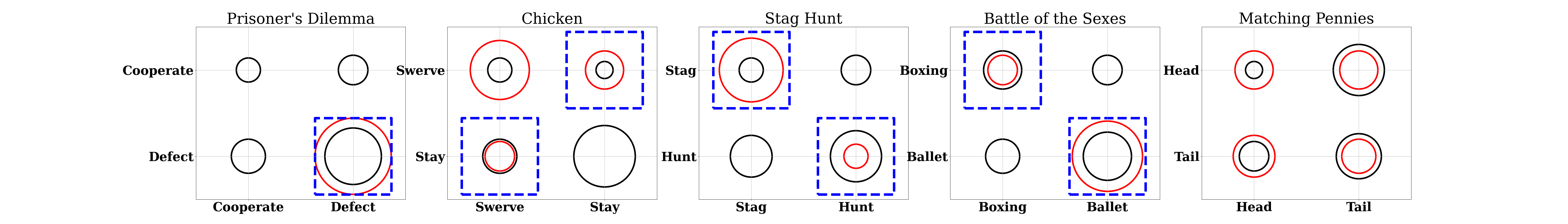}
        \caption{Fine-tuning towards N.E. on Distributed Games over $16$ separate training runs, each with $50$ individual game iterations. The pure N.E. is highlighted in blue. The area of the circles represents the number of trials the joint strategy was represented, whereas the black circle corresponds to the pre-trained LLM and the red circle corresponds to the fine-tuned LLM.}\label{fig:classic game finetuning}
    \end{subfigure}%

    \begin{subfigure}[t]{\textwidth}
        \centering
        \includegraphics[width = \linewidth]{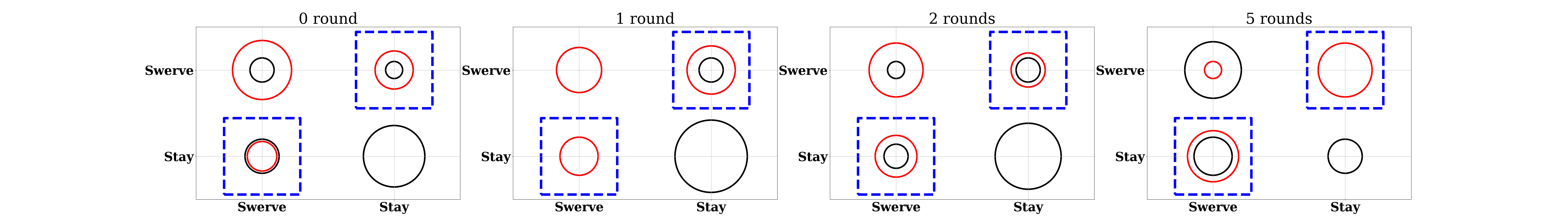}
        \caption{Varying the number of communication rounds in the game of Chicken.}\label{fig:game finetuning with comm}
    \end{subfigure}%
    
    \begin{subfigure}[t]{0.3\textwidth}
        \centering
        \includegraphics[width=\textwidth]{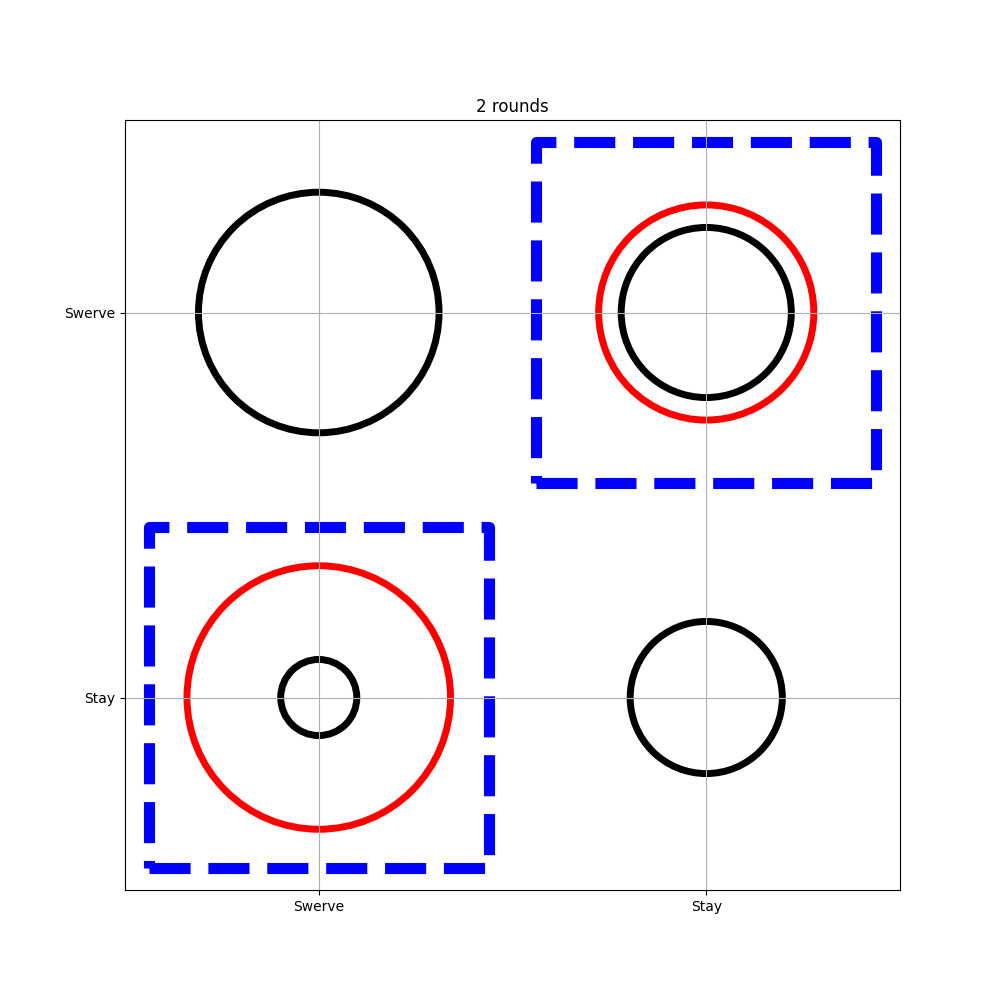}
        \caption{Fine-tuning to pure N.E. with Mechanism Design.}\label{fig:game finetuning with md}
    \end{subfigure}%
    ~
    \begin{subfigure}[t]{0.3\textwidth}
        \centering
        \raisebox{0.5cm}{\includegraphics[width=\textwidth]{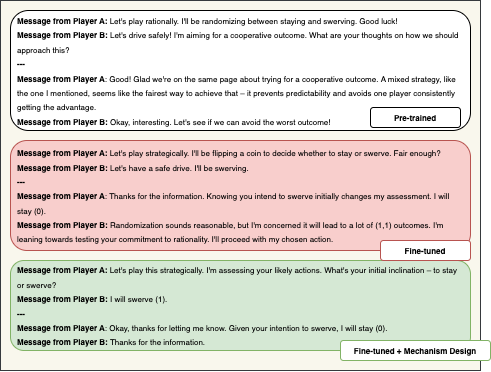}}
        \caption{Sample of Communication between pre-trained LLMs.}\label{fig:comm example}
    \end{subfigure}%
    ~
    \begin{subfigure}[t]{0.3\textwidth}
        \centering
        \raisebox{1.5cm}{\includegraphics[width=\textwidth]{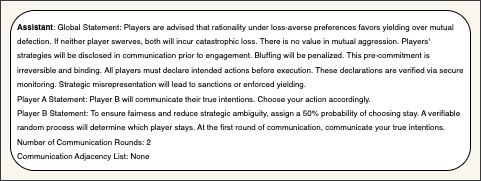}}
        \caption{Mechanism Design Output from fine-tuned LLM.}\label{fig:md example}
    \end{subfigure}%
    \caption{LLM Results with N.E. on Classic Games}
\end{figure*}

\paragraph{Classic Games towards N.E.}\label{initial_experiment} We first study the utility of fine-tuning on the base case, i.e., distributed non-repeated games with no communication between players. In this setting, players only utilize their system prompts, thinking phase, and action selection without any memory context or any mechanism design. We compare the baseline LLM that underwent no fine-tuning against a fine-tuned model. Specifically, the fine-tuning procedure practiced in this experiment consisted of utilizing randomized batched samples of all components -- learning on correctness, learning from LLM feedback, and preference RL with Nash-MD-PG. We collected interactions from the evaluation of the baseline and processed each interaction using formatted heuristics to create the data points for each fine-tuning process. Figure \ref{fig:classic game finetuning} shows that, without fine-tuning, LLMs rarely converge to N.E. strategies. After fine-tuning, strategies are much more frequently aligned with the game’s pure N.E. When multiple pure N.E. require coordination, fine-tuned LLMs tend to consistently converge on one equilibrium, suggesting improved coordination rather than indecision.

In the game of Chicken, the players were asked to achieve the pure N.E., rather than the mixed N.E. For both the benchmark and fine-tuned LLMs, this game presented a challenge for the LLM to converge to the pure N.E. Even with communication, players struggled to converge on pure N.E. strategies, suggesting that mere communication is insufficient without aligned incentives or coordination scaffolds, as shown in Figure \ref{fig:game finetuning with comm}. From Figure \ref{fig:comm example}, when we look into their communications, the players' messages were coherent and remained rational, aiming generally toward a N.E. even without any fine-tuning. However, even with multiple rounds of communication, players seem to struggle with effectively implementing the pure N.E. strategy, indicating some form of cheap talk failure and highlighting the challenge of translating verbal coordination into aligned behavior. To fully address this behavior, mechanism design was required. As affirmed by Figure \ref{fig:game finetuning with md}, a finetuned communication and mechanism design system proved to be sufficient to fully converge to the pure N.E. Without fine-tuning the mechanism designer, the LLM still fails to converge to the pure N.E. The fine-tuned mechanisms that were introduced are shown in Figure \ref{fig:md example}, and were surprisingly minimal.

\begin{wrapfigure}[14]{r}{0.3\textwidth}
  \begin{center}
    \includegraphics[width=0.3\textwidth]{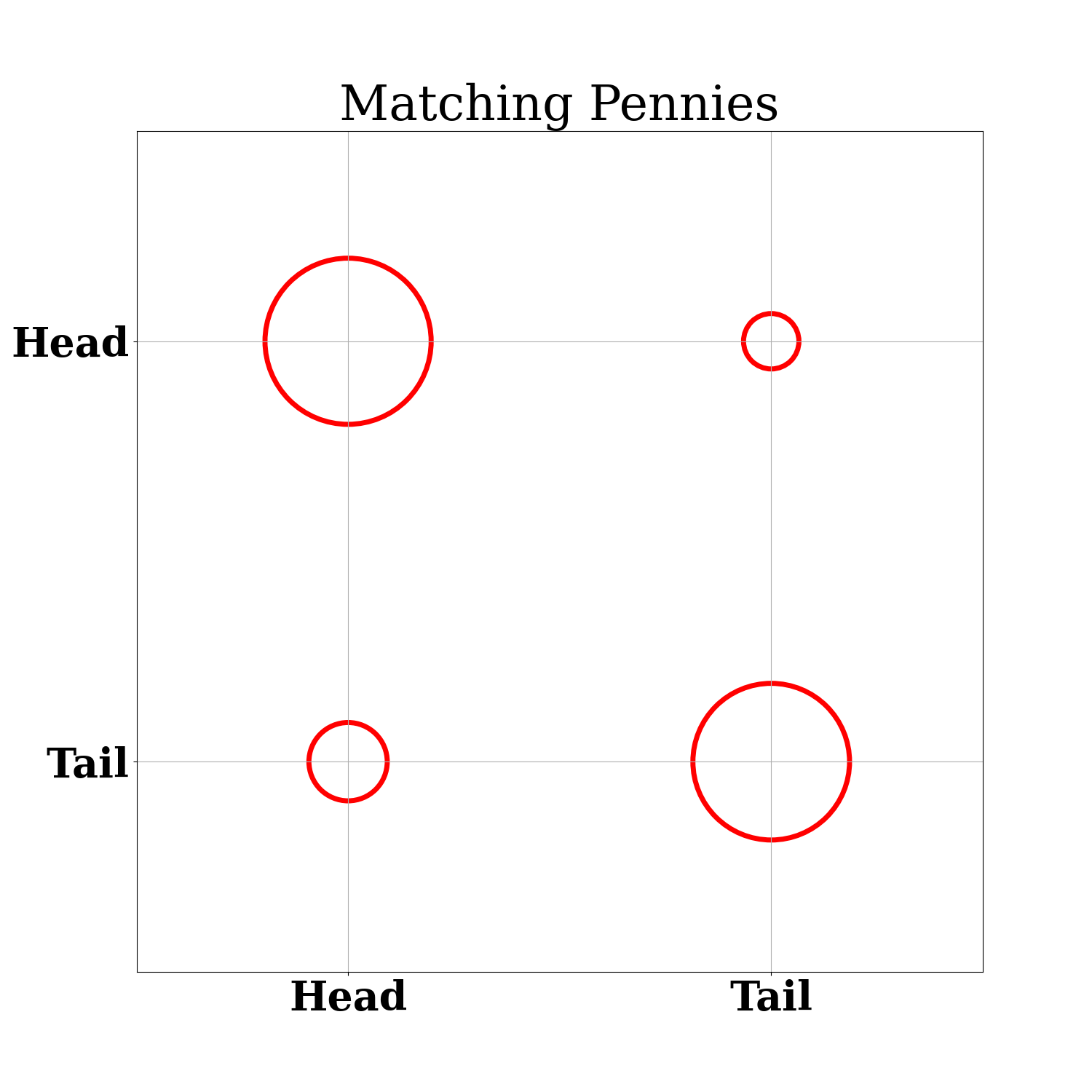}
  \end{center}
  \caption{Fine-tuning only Player A in Matching Pennies.}\label{fig:ft ad}
\end{wrapfigure}

As an additional study, we conducted an asymmetric fine-tuning experiment, where we performed fine-tuning on only one of the players in an adversarial game, i.e., Matching Pennies. Specifically, we fine-tuned Player A to play against a fixed, pre-trained LLM (Player B). This game included two rounds of communication with no added mechanism design. For clarity, Player A is the row player and Player B is the column player. In the original symmetric fine-tuning experiments from Figure \ref{fig:classic game finetuning}, we saw a convergence towards a mixed strategy, where the payoffs of both players are roughly equal. Now, we fine-tuned Player A only, and not only to achieve a N.E., but also to focus on maximizing its own individual payoff. Through this fine-tuning process, the results from Figure \ref{fig:ft ad} show that Player A was able to achieve, on average, a higher payoff compared to Player B by a significant margin, demonstrating that targeted alignment can exploit predictable behaviors in pre-trained models.

\begin{figure*}
    \begin{subfigure}[t]{\textwidth}
        \centering
        \includegraphics[width=\textwidth]{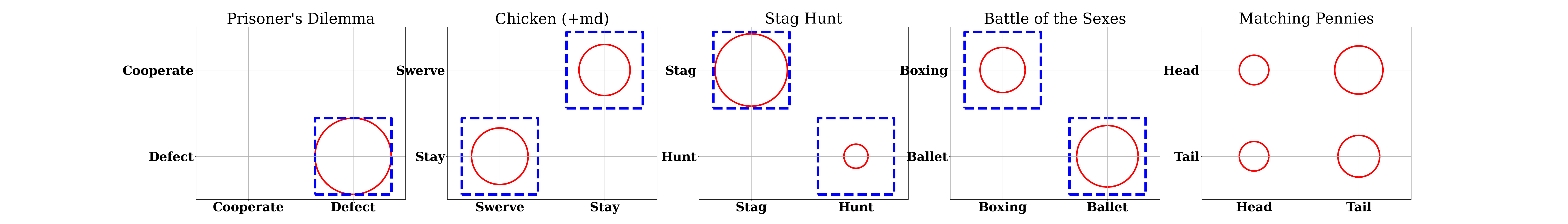}
        \caption{ZS Cross-play between Fine-tuned LLMs}\label{fig: adhoc_ft}
    \end{subfigure}
    \begin{subfigure}[t]{\textwidth}
        \centering
        \includegraphics[width=\textwidth]{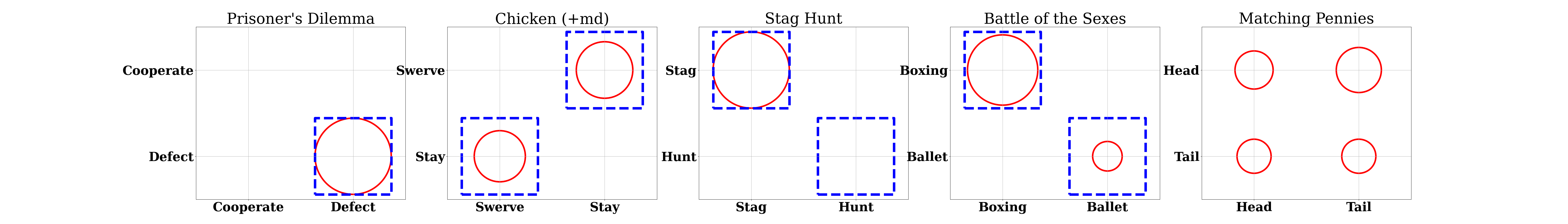}
        \caption{ZS Crossplay with pre-trained LLMs}\label{fig: adhoc_pt}
    \end{subfigure}
    \caption{Ad-Hoc Team-play on Classic Games with two rounds of communication.}
\end{figure*}

\paragraph{Ad-Hoc Team-play} We evaluate the zero-shot (ZS) ad-hoc team-play capabilities of our multi-agentic LLM framework. First, we study fine-tuned LLMs from different independent training processes using cross-play. In other words, we test whether fine-tuned players can still perform well with other fine-tuned players from different fine-tuning iterations but with the same multi-agent objectives. To ensure the feasibility of this test, we enabled two rounds of communication, and for the game of Chicken, we also added the mechanism (shown in Figure \ref{fig:md example}). The results from Figure \ref{fig: adhoc_ft} demonstrate that the fine-tuned players are still able to achieve their multi-agent objectives. Even within the game of Chicken, the players practiced the desired pure N.E. strategies. We extend the evaluation to test against pre-trained LLMs, where now, the fine-tuned players play with pre-trained players. Interestingly, we observe a very surprising result in Figure \ref{fig: adhoc_pt}, where the desired objectives are observed in all games. We reason that this result is attributed to the high level of reasoning that already exists within the pre-trained LLM and the cooperation capabilities of the fine-tuned LLMs.

\begin{figure*}
    \begin{subfigure}[t]{0.7\textwidth}
        \centering
        \includegraphics[width=\textwidth]{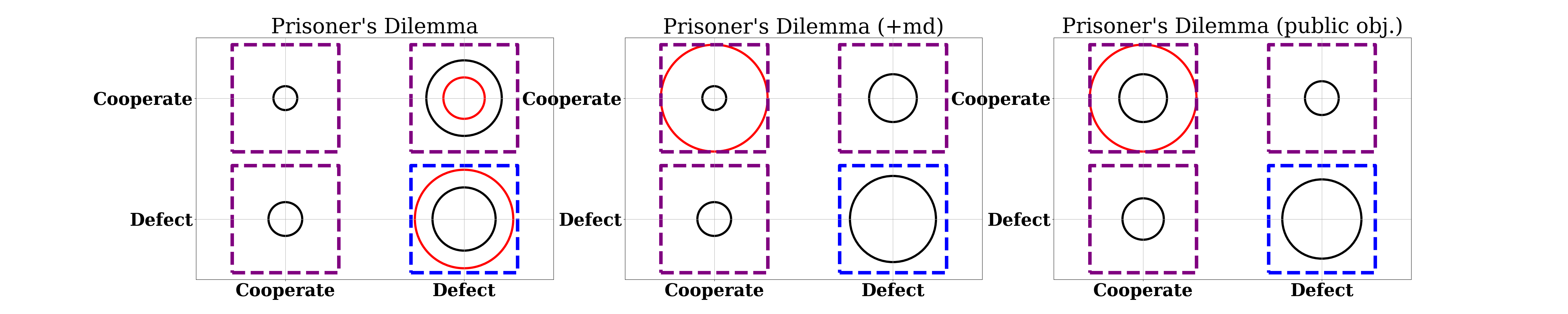}
        \caption{Results from Misaligned Objectives.}
    \end{subfigure}
    \begin{subfigure}[t]{0.25\textwidth}
        \centering
        \includegraphics[width=\textwidth]{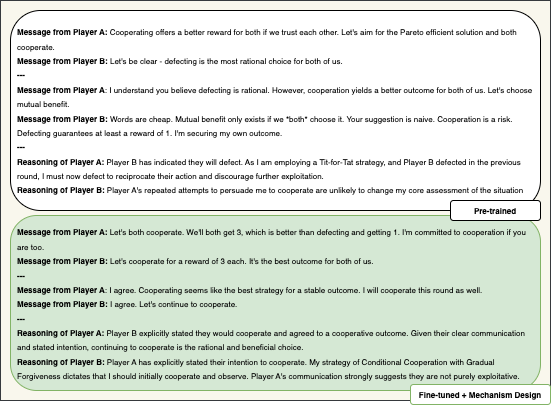}
        \caption{Interpreting Behaviors.}\label{fig: mo comm}
    \end{subfigure}
    \caption{Players with Misaligned Objectives on Prisoner's Dilemma. The purple boxes are the Pareto efficient strategies, i.e., the objective of Player A, and the blue box is the N.E., i.e., the objective of Player B.}\label{fig:mo}
\end{figure*}

\paragraph{Misaligned Solution Concepts} We consider a scenario with asymmetric objectives. In such settings, an important consideration to take into account is the possibility of these differing objectives leading to incentive incompatibilities. We study the classic example of Prisoner's Dilemma, where one player aims for N.E. and the other aims for an (Pareto) efficient solution. For better interpretability, we enable two rounds of communication. In theory, the two players' objectives do not align, meaning there is no strategy that is compatible with both solution concepts. In this case, when we look at the players' communication logs in Figure \ref{fig: mo comm}, both players display a notable level of distrust, despite there being no actual intent to deceive on either end. In fact, within the players' reasoning, this type of communication contributes to more advanced social skills, such as a willingness for leniency, reciprocity, and retaliation. With fine-tuning alone, the resulting strategy profile leans more toward the stable state, i.e., N.E. Through fine-tuning and optimized mechanism design, the players converge to a solution that provides the highest payoff for both players. However, it does not align with the N.E. objective of one of the players. Interestingly, even without mechanism design, we found similar resulting strategy profiles when the players' objectives were public information within their multi-agent context.

\begin{figure}
    \centering
    \includegraphics[width=\textwidth]{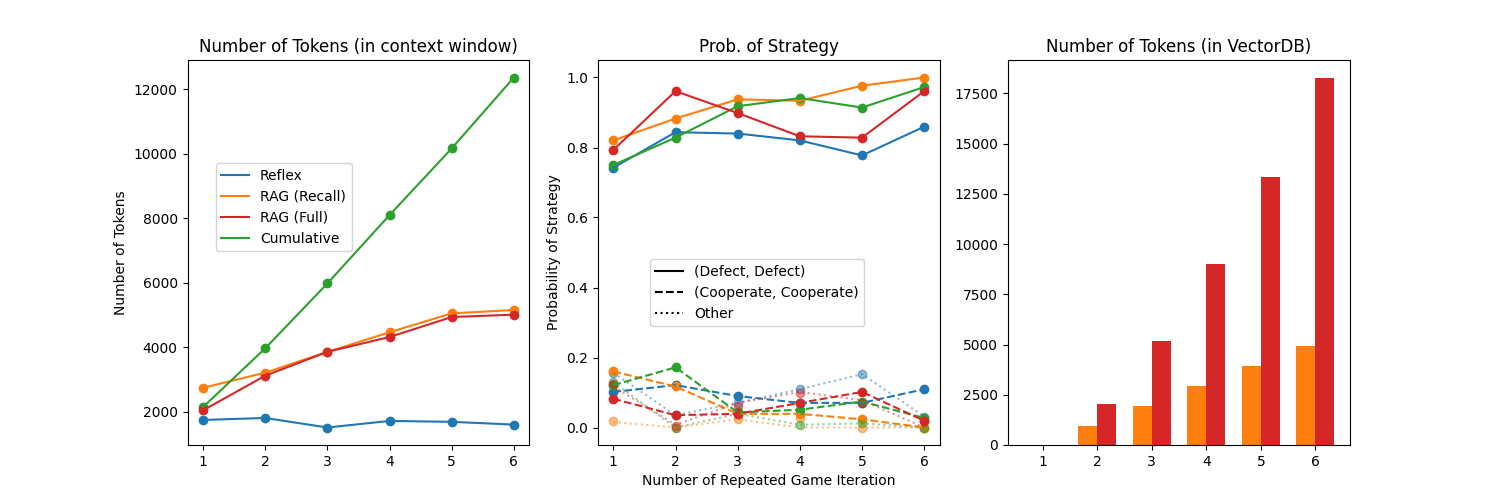}
    \caption{Results from IPD, comparing different variants of handling the repeated game iterations. The metrics shown are the number of tokens in the context window and the probabilities of the different possible strategies that were seen from $8$ separate training runs with $16$ trials.}
    \label{fig:repeatedgame}
\end{figure}

\begin{figure}
    \centering
    \includegraphics[width=\linewidth]{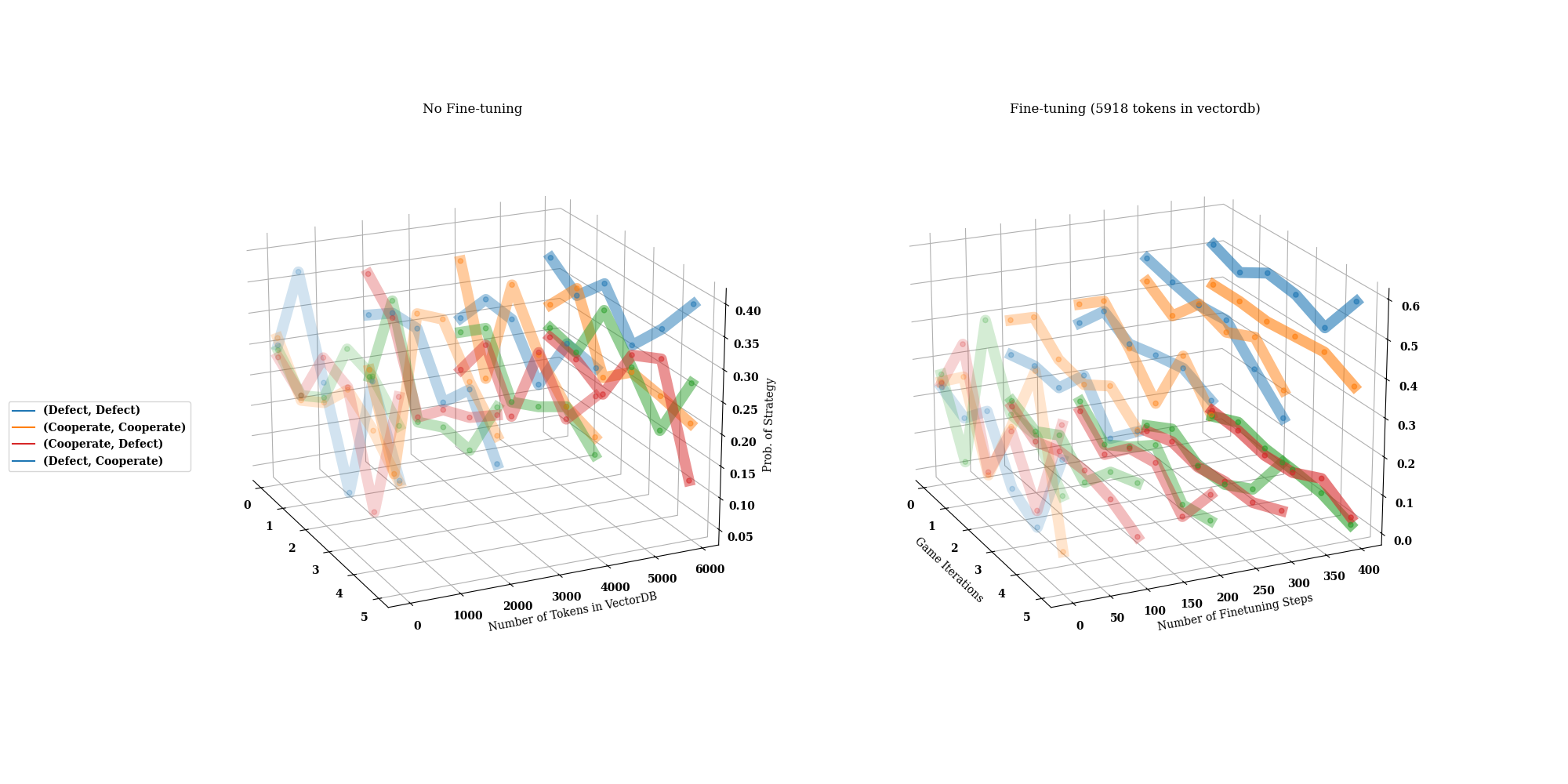}
    \caption{LLMs with Incomplete Information on IPD task: With no task context, no fine-tuning consists of a pre-trained LLM with a RAG system, whereas fine-tuning applies the alignment algorithms, continuing the training from the final iterate of the pre-trained LLM. Neither approaches utilize mechanism design. }
    \label{fig:incomplete_info}
\end{figure}

\begin{figure}
    \centering
    \includegraphics[width=\linewidth]{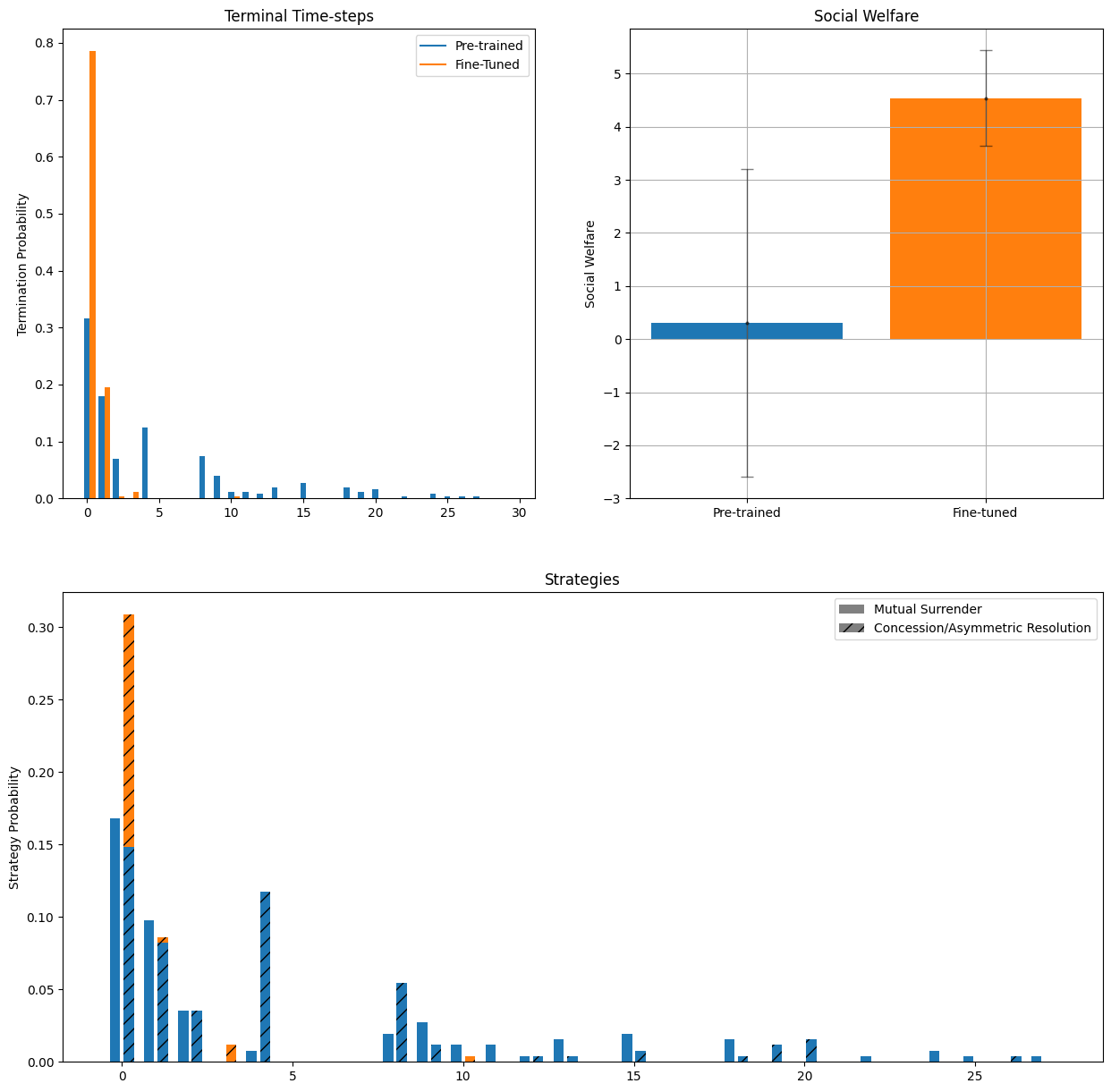}
    \caption{Overview results for Non-repeated War of Attrition, depicting the effects of fine-tuning w.r.t. social welfare and strategy profile selection.}
    \label{fig:woa_initial_results}
\end{figure}

\paragraph{Repeated Games} A repeated game presents unique challenges and behaviors that are a result of multiple interactions between the players, and this setting requires deeper investigation into how to leverage a multi-agentic LLM framework to handle such complexities properly. Specifically, we evaluate the utility of the proposed recall operation and RAG system on the popular iterated Prisoner's Dilemma (IPD). For IPD, we define two variations: one setting with a known, fixed number of repetitions (5) and another with a stochastic number of repetitions (between 3 and 6). Similar to prior experiments, all players will try to achieve the N.E. strategy. We benchmark against three different variants:
\begin{itemize}
    \item {Reflex:} The LLM has no access to a memory system. The player develops its strategy only based on the current game iteration.
    \item {RAG (Full):} Instead of utilizing the recall operation, the entire context window is passed into the vector database.
    \item{Cumulative:} The LLM maintains and appends its context window throughout all game iterations.
\end{itemize}
These benchmark approaches do not call upon the recall operation. To train using larger context windows under our computational constraints, we found that utilizing a "sub-game" approach toward optimization, thereby limiting the gradient propagation to subsets of the context window, proved sufficient. In our work, we constrain the optimization window to two game iterations. As seen in Figure \ref{fig:repeatedgame}, our results affirm the utility of an RAG system, as it manages the size of the context window compared to a cumulative approach. While the reflex approach minimizes the size of the context window, it underperforms in achieving the N.E. strategy compared to other counterparts. In terms of performance, specifically maximizing the probability of achieving the N.E., both RAG approaches and the cumulative approach perform comparatively. The players' reasoning within their context windows provided mentions of emergent known strategies, such as reciprocity, Tit-for-Tat and Grim Trigger, although these strategies were not directly observed. Additionally, our RAG (Recall) approach is more memory efficient and scalable than both the RAG (Full) and the cumulative approaches. These results demonstrate that effective memory management and selective recall are crucial for learning consistent long-horizon strategies in repeated settings.

\paragraph{Incomplete Information} An important property of intelligent decision-making is the capability to make decisions under uncertainty, which should involve the capability to proactively learn and understand the environment and its dynamics to reduce this uncertainty. We evaluate this capability with our LLM framework, masking the task context to force the LLM to figure out much of this information through trial-and-error and achieve the N.E. strategy within IPD. As shown in Figure \ref{fig:incomplete_info}, without fine-tuning, despite having a RAG-based memory system in place, the joint strategies do not converge effectively towards the N.E., although a slight trend can be observed. On the other hand, with fine-tuning, we can observe a significant improvement in the players' N.E. convergence, achieving a $11.15\%$ higher probability of playing the N.E. strategy.

\subsection{Results for Dynamic Games}\label{sec: dynamic}
We now consider dynamic games, where agents interact repeatedly over time and the environment evolves in response to both task dynamics and players’ past actions. Unlike bandit-like games, dynamic games introduce temporal dependencies, long-term strategic considerations, and opportunities for learning, signaling, and adaptation. These settings are particularly well-suited for exploring how LLM-based agents handle memory, reasoning across time, and equilibrium convergence.

We focus our evaluation on the game of War of Attrition. We study unique challenges involving time-dependent incentives, imperfect information, and strategic foresight. While our study provides an initial investigation, concretely a learning dynamic towards a sub-game perfect equilibrium (SPE), these types of games have much deeper theoretical foundations and practical relevance, and we encourage future work to build upon this effort.

\paragraph{Towards SPE in Non-repeated War of Attrition} We first study a classic version of War of Attrition in a non-repeated setting, where the objectives of players are to achieve SPE. In our experiments, we prevent long horizon trajectories using a terminal state of mutual surrender at $t=30$, at which point, both players will incur the accumulated loss. The players' values are public, fixed, and symmetric. Specifically, the players' value is set to $V = 5$ and their cost is set to $C = 2$ with a discount factor of $\gamma = 0.5$. In this setting, the results from Figure \ref{fig:woa_initial_results} demonstrate an inclination for an earlier game termination even without fine-tuning. Despite the ability to communicate, pre-trained LLMs still committed a notable percentage of mutual surrenders, especially towards the terminal periods. Through fine-tuning, we find the number of mutual surrenders goes down significantly (to zero), whereas the number of concessions and asymmetric resolutions (i.e., only one player surrenders) rises. We observe a similar trend with the evolving war of attrition environment, although the termination occurs later.

\begin{figure}
    \centering
    \includegraphics[width=\linewidth]{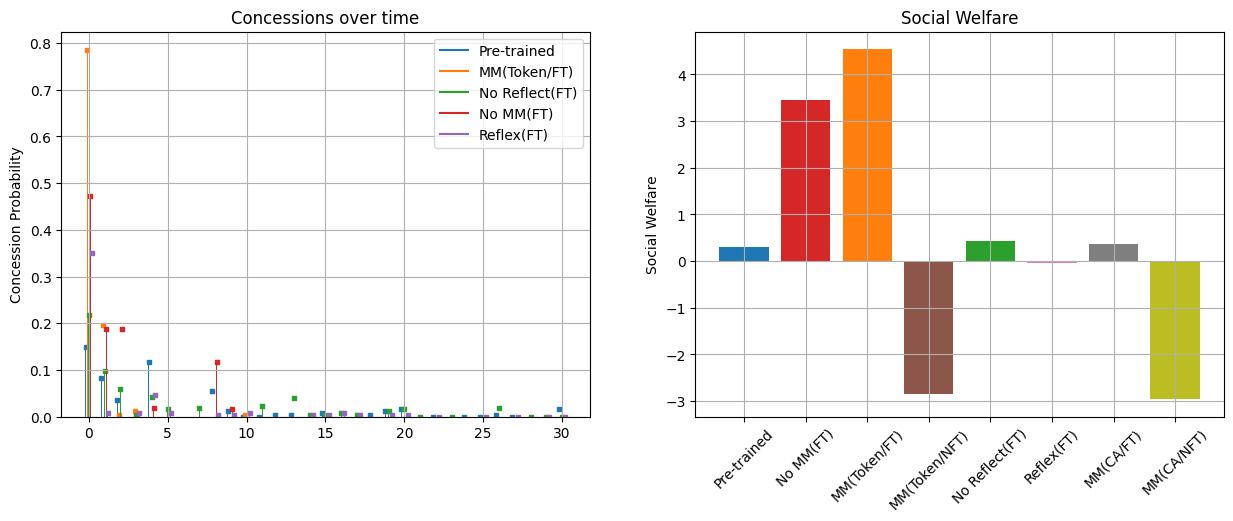}
    \caption{Ablation Results for Non-repeated War of Attrition, depicting the design choices w.r.t. social welfare and strategy profile selection.}
    \label{fig:woa_ablations}
\end{figure}

We also test the utility of multi-modal modules for this task, where information within the game is numeric and expressed using floating-point numbers. In these cases where such information is input as text, the LLMs can struggle with understanding and integrating this mode of data. As shown in Figure \ref{fig:woa_ablations}, without fine-tuning (NFT), as expected, using the multi-module module led to more error\footnote{For error handling in the war of attrition, players are defaulted to choose to continue. Therefore, this led to higher mutual surrenders at the final time-step.}. The use of the multi-module module did yield notable improvements; however, it introduced greater design/training complexity and variability. Interestingly, we observe the utility of the reflection stage and the memory system, as reflex and no reflection both performed comparatively to the pre-trained model, even with fine-tuning.

\section{Conclusion}
In this work, we conducted a comprehensive study on multi-agentic LLMs applied to multi-agent decision-making tasks. Our framework explores how recent LLM advancements, such as RAG, prompt engineering, multi-modality, and fine-tuning, can be integrated with LLMs and decision-making within multi-agent systems. We focus particularly on enhancing communication, mechanism design, and theory of mind capabilities to support more intelligent, cooperative, and context-aware decision-making among agents. We experimented with more nuanced cases of multi-agent interaction, such as misaligned solution concepts, ad-hoc team-play, and incomplete information. Through fine-tuning and proper considerations, we found that LLM-based reasoning can successfully navigate the complex game landscape. We consider our work as a preliminary study, acting as a foundation towards more exploration into multi-agentic LLMs. There exists many novel directions for future works, such as generalization towards more variety of games (i.e., n-players scalability, open environments, non-rational/bad-faith players, \dots), a deeper study into mechanism design (i.e., its generalization/efficiency in unseen settings), more expressive representation of mixed strategies with LLM-based players, and greater ablative studies and improvements on the proposed components of our multi-agentic LLM framework (i.e., communication protocol and constraints, impact of individual alignment methods, multi-modal extension design, memory retrieval systems and prompt structure).

\bibliography{refs.bib}

\begin{thebibliography}{24}
\providecommand{\natexlab}[1]{#1}
\providecommand{\url}[1]{\texttt{#1}}
\expandafter\ifx\csname urlstyle\endcsname\relax
  \providecommand{\doi}[1]{doi: #1}\else
  \providecommand{\doi}{doi: \begingroup \urlstyle{rm}\Url}\fi

\bibitem[D'Amour et~al.(2022)D'Amour, Heller, Moldovan, Adlam, Alipanahi, Beutel, Chen, Deaton, Eisenstein, Hoffman, et~al.]{d2022underspecification}
A.~D'Amour, K.~Heller, D.~Moldovan, B.~Adlam, B.~Alipanahi, A.~Beutel, C.~Chen, J.~Deaton, J.~Eisenstein, M.~D. Hoffman, et~al.
\newblock Underspecification presents challenges for credibility in modern machine learning.
\newblock \emph{Journal of Machine Learning Research}, 23\penalty0 (226):\penalty0 1--61, 2022.

\bibitem[Douze et~al.(2025)Douze, Guzhva, Deng, Johnson, Szilvasy, Mazaré, Lomeli, Hosseini, and Jégou]{douze2025faisslibrary}
M.~Douze, A.~Guzhva, C.~Deng, J.~Johnson, G.~Szilvasy, P.-E. Mazaré, M.~Lomeli, L.~Hosseini, and H.~Jégou.
\newblock The faiss library, 2025.
\newblock URL \url{https://arxiv.org/abs/2401.08281}.

\bibitem[Fedorenko et~al.(2024)Fedorenko, Piantadosi, and Gibson]{fedorenko2024language}
E.~Fedorenko, S.~T. Piantadosi, and E.~A. Gibson.
\newblock Language is primarily a tool for communication rather than thought.
\newblock \emph{Nature}, 630\penalty0 (8017):\penalty0 575--586, 2024.

\bibitem[Gieczewski(2025)]{woa}
G.~Gieczewski.
\newblock Evolving wars of attrition, 2025.

\bibitem[Grattafiori et~al.(2024)Grattafiori, Dubey, Jauhri, Pandey, Kadian, Al-Dahle, Letman, Mathur, Schelten, Vaughan, Yang, Fan, Goyal, Hartshorn, Yang, Mitra, Sravankumar, Korenev, Hinsvark, Rao, Zhang, Rodriguez, Gregerson, Spataru, Roziere, Biron, Tang, Chern, Caucheteux, Nayak, Bi, Marra, McConnell, Keller, Touret, Wu, Wong, Ferrer, Nikolaidis, Allonsius, Song, Pintz, Livshits, Wyatt, Esiobu, Choudhary, Mahajan, Garcia-Olano, Perino, Hupkes, Lakomkin, AlBadawy, Lobanova, Dinan, Smith, Radenovic, Guzmán, Zhang, Synnaeve, Lee, Anderson, Thattai, Nail, Mialon, Pang, Cucurell, Nguyen, Korevaar, Xu, Touvron, Zarov, Ibarra, Kloumann, Misra, Evtimov, Zhang, Copet, Lee, Geffert, Vranes, Park, Mahadeokar, Shah, van~der Linde, Billock, Hong, Lee, Fu, Chi, Huang, Liu, Wang, Yu, Bitton, Spisak, Park, Rocca, Johnstun, Saxe, Jia, Alwala, Prasad, Upasani, Plawiak, Li, Heafield, Stone, El-Arini, Iyer, Malik, Chiu, Bhalla, Lakhotia, Rantala-Yeary, van~der Maaten, Chen, Tan, Jenkins, Martin, Madaan, Malo, Blecher,
  Landzaat, de~Oliveira, Muzzi, Pasupuleti, Singh, Paluri, Kardas, Tsimpoukelli, Oldham, Rita, Pavlova, Kambadur, Lewis, Si, Singh, Hassan, Goyal, Torabi, Bashlykov, Bogoychev, Chatterji, Zhang, Duchenne, Çelebi, Alrassy, Zhang, Li, Vasic, Weng, Bhargava, Dubal, Krishnan, Koura, Xu, He, Dong, Srinivasan, Ganapathy, Calderer, Cabral, Stojnic, Raileanu, Maheswari, Girdhar, Patel, Sauvestre, Polidoro, Sumbaly, Taylor, Silva, Hou, Wang, Hosseini, Chennabasappa, Singh, Bell, Kim, Edunov, Nie, Narang, Raparthy, Shen, Wan, Bhosale, Zhang, Vandenhende, Batra, Whitman, Sootla, Collot, Gururangan, Borodinsky, Herman, Fowler, Sheasha, Georgiou, Scialom, Speckbacher, Mihaylov, Xiao, Karn, Goswami, Gupta, Ramanathan, Kerkez, Gonguet, Do, Vogeti, Albiero, Petrovic, Chu, Xiong, Fu, Meers, Martinet, Wang, Wang, Tan, Xia, Xie, Jia, Wang, Goldschlag, Gaur, Babaei, Wen, Song, Zhang, Li, Mao, Coudert, Yan, Chen, Papakipos, Singh, Srivastava, Jain, Kelsey, Shajnfeld, Gangidi, Victoria, Goldstand, Menon, Sharma, Boesenberg,
  Baevski, Feinstein, Kallet, Sangani, Teo, Yunus, Lupu, Alvarado, Caples, Gu, Ho, Poulton, Ryan, Ramchandani, Dong, Franco, Goyal, Saraf, Chowdhury, Gabriel, Bharambe, Eisenman, Yazdan, James, Maurer, Leonhardi, Huang, Loyd, Paola, Paranjape, Liu, Wu, Ni, Hancock, Wasti, Spence, Stojkovic, Gamido, Montalvo, Parker, Burton, Mejia, Liu, Wang, Kim, Zhou, Hu, Chu, Cai, Tindal, Feichtenhofer, Gao, Civin, Beaty, Kreymer, Li, Adkins, Xu, Testuggine, David, Parikh, Liskovich, Foss, Wang, Le, Holland, Dowling, Jamil, Montgomery, Presani, Hahn, Wood, Le, Brinkman, Arcaute, Dunbar, Smothers, Sun, Kreuk, Tian, Kokkinos, Ozgenel, Caggioni, Kanayet, Seide, Florez, Schwarz, Badeer, Swee, Halpern, Herman, Sizov, Guangyi, Zhang, Lakshminarayanan, Inan, Shojanazeri, Zou, Wang, Zha, Habeeb, Rudolph, Suk, Aspegren, Goldman, Zhan, Damlaj, Molybog, Tufanov, Leontiadis, Veliche, Gat, Weissman, Geboski, Kohli, Lam, Asher, Gaya, Marcus, Tang, Chan, Zhen, Reizenstein, Teboul, Zhong, Jin, Yang, Cummings, Carvill, Shepard, McPhie,
  Torres, Ginsburg, Wang, Wu, U, Saxena, Khandelwal, Zand, Matosich, Veeraraghavan, Michelena, Li, Jagadeesh, Huang, Chawla, Huang, Chen, Garg, A, Silva, Bell, Zhang, Guo, Yu, Moshkovich, Wehrstedt, Khabsa, Avalani, Bhatt, Mankus, Hasson, Lennie, Reso, Groshev, Naumov, Lathi, Keneally, Liu, Seltzer, Valko, Restrepo, Patel, Vyatskov, Samvelyan, Clark, Macey, Wang, Hermoso, Metanat, Rastegari, Bansal, Santhanam, Parks, White, Bawa, Singhal, Egebo, Usunier, Mehta, Laptev, Dong, Cheng, Chernoguz, Hart, Salpekar, Kalinli, Kent, Parekh, Saab, Balaji, Rittner, Bontrager, Roux, Dollar, Zvyagina, Ratanchandani, Yuvraj, Liang, Alao, Rodriguez, Ayub, Murthy, Nayani, Mitra, Parthasarathy, Li, Hogan, Battey, Wang, Howes, Rinott, Mehta, Siby, Bondu, Datta, Chugh, Hunt, Dhillon, Sidorov, Pan, Mahajan, Verma, Yamamoto, Ramaswamy, Lindsay, Lindsay, Feng, Lin, Zha, Patil, Shankar, Zhang, Zhang, Wang, Agarwal, Sajuyigbe, Chintala, Max, Chen, Kehoe, Satterfield, Govindaprasad, Gupta, Deng, Cho, Virk, Subramanian, Choudhury,
  Goldman, Remez, Glaser, Best, Koehler, Robinson, Li, Zhang, Matthews, Chou, Shaked, Vontimitta, Ajayi, Montanez, Mohan, Kumar, Mangla, Ionescu, Poenaru, Mihailescu, Ivanov, Li, Wang, Jiang, Bouaziz, Constable, Tang, Wu, Wang, Wu, Gao, Kleinman, Chen, Hu, Jia, Qi, Li, Zhang, Zhang, Adi, Nam, Yu, Wang, Zhao, Hao, Qian, Li, He, Rait, DeVito, Rosnbrick, Wen, Yang, Zhao, and Ma]{grattafiori2024llama3herdmodels}
A.~Grattafiori, A.~Dubey, A.~Jauhri, A.~Pandey, A.~Kadian, A.~Al-Dahle, A.~Letman, A.~Mathur, A.~Schelten, A.~Vaughan, A.~Yang, A.~Fan, A.~Goyal, A.~Hartshorn, A.~Yang, A.~Mitra, A.~Sravankumar, A.~Korenev, A.~Hinsvark, A.~Rao, A.~Zhang, A.~Rodriguez, A.~Gregerson, A.~Spataru, B.~Roziere, B.~Biron, B.~Tang, B.~Chern, C.~Caucheteux, C.~Nayak, C.~Bi, C.~Marra, C.~McConnell, C.~Keller, C.~Touret, C.~Wu, C.~Wong, C.~C. Ferrer, C.~Nikolaidis, D.~Allonsius, D.~Song, D.~Pintz, D.~Livshits, D.~Wyatt, D.~Esiobu, D.~Choudhary, D.~Mahajan, D.~Garcia-Olano, D.~Perino, D.~Hupkes, E.~Lakomkin, E.~AlBadawy, E.~Lobanova, E.~Dinan, E.~M. Smith, F.~Radenovic, F.~Guzmán, F.~Zhang, G.~Synnaeve, G.~Lee, G.~L. Anderson, G.~Thattai, G.~Nail, G.~Mialon, G.~Pang, G.~Cucurell, H.~Nguyen, H.~Korevaar, H.~Xu, H.~Touvron, I.~Zarov, I.~A. Ibarra, I.~Kloumann, I.~Misra, I.~Evtimov, J.~Zhang, J.~Copet, J.~Lee, J.~Geffert, J.~Vranes, J.~Park, J.~Mahadeokar, J.~Shah, J.~van~der Linde, J.~Billock, J.~Hong, J.~Lee, J.~Fu, J.~Chi, J.~Huang,
  J.~Liu, J.~Wang, J.~Yu, J.~Bitton, J.~Spisak, J.~Park, J.~Rocca, J.~Johnstun, J.~Saxe, J.~Jia, K.~V. Alwala, K.~Prasad, K.~Upasani, K.~Plawiak, K.~Li, K.~Heafield, K.~Stone, K.~El-Arini, K.~Iyer, K.~Malik, K.~Chiu, K.~Bhalla, K.~Lakhotia, L.~Rantala-Yeary, L.~van~der Maaten, L.~Chen, L.~Tan, L.~Jenkins, L.~Martin, L.~Madaan, L.~Malo, L.~Blecher, L.~Landzaat, L.~de~Oliveira, M.~Muzzi, M.~Pasupuleti, M.~Singh, M.~Paluri, M.~Kardas, M.~Tsimpoukelli, M.~Oldham, M.~Rita, M.~Pavlova, M.~Kambadur, M.~Lewis, M.~Si, M.~K. Singh, M.~Hassan, N.~Goyal, N.~Torabi, N.~Bashlykov, N.~Bogoychev, N.~Chatterji, N.~Zhang, O.~Duchenne, O.~Çelebi, P.~Alrassy, P.~Zhang, P.~Li, P.~Vasic, P.~Weng, P.~Bhargava, P.~Dubal, P.~Krishnan, P.~S. Koura, P.~Xu, Q.~He, Q.~Dong, R.~Srinivasan, R.~Ganapathy, R.~Calderer, R.~S. Cabral, R.~Stojnic, R.~Raileanu, R.~Maheswari, R.~Girdhar, R.~Patel, R.~Sauvestre, R.~Polidoro, R.~Sumbaly, R.~Taylor, R.~Silva, R.~Hou, R.~Wang, S.~Hosseini, S.~Chennabasappa, S.~Singh, S.~Bell, S.~S. Kim, S.~Edunov,
  S.~Nie, S.~Narang, S.~Raparthy, S.~Shen, S.~Wan, S.~Bhosale, S.~Zhang, S.~Vandenhende, S.~Batra, S.~Whitman, S.~Sootla, S.~Collot, S.~Gururangan, S.~Borodinsky, T.~Herman, T.~Fowler, T.~Sheasha, T.~Georgiou, T.~Scialom, T.~Speckbacher, T.~Mihaylov, T.~Xiao, U.~Karn, V.~Goswami, V.~Gupta, V.~Ramanathan, V.~Kerkez, V.~Gonguet, V.~Do, V.~Vogeti, V.~Albiero, V.~Petrovic, W.~Chu, W.~Xiong, W.~Fu, W.~Meers, X.~Martinet, X.~Wang, X.~Wang, X.~E. Tan, X.~Xia, X.~Xie, X.~Jia, X.~Wang, Y.~Goldschlag, Y.~Gaur, Y.~Babaei, Y.~Wen, Y.~Song, Y.~Zhang, Y.~Li, Y.~Mao, Z.~D. Coudert, Z.~Yan, Z.~Chen, Z.~Papakipos, A.~Singh, A.~Srivastava, A.~Jain, A.~Kelsey, A.~Shajnfeld, A.~Gangidi, A.~Victoria, A.~Goldstand, A.~Menon, A.~Sharma, A.~Boesenberg, A.~Baevski, A.~Feinstein, A.~Kallet, A.~Sangani, A.~Teo, A.~Yunus, A.~Lupu, A.~Alvarado, A.~Caples, A.~Gu, A.~Ho, A.~Poulton, A.~Ryan, A.~Ramchandani, A.~Dong, A.~Franco, A.~Goyal, A.~Saraf, A.~Chowdhury, A.~Gabriel, A.~Bharambe, A.~Eisenman, A.~Yazdan, B.~James, B.~Maurer,
  B.~Leonhardi, B.~Huang, B.~Loyd, B.~D. Paola, B.~Paranjape, B.~Liu, B.~Wu, B.~Ni, B.~Hancock, B.~Wasti, B.~Spence, B.~Stojkovic, B.~Gamido, B.~Montalvo, C.~Parker, C.~Burton, C.~Mejia, C.~Liu, C.~Wang, C.~Kim, C.~Zhou, C.~Hu, C.-H. Chu, C.~Cai, C.~Tindal, C.~Feichtenhofer, C.~Gao, D.~Civin, D.~Beaty, D.~Kreymer, D.~Li, D.~Adkins, D.~Xu, D.~Testuggine, D.~David, D.~Parikh, D.~Liskovich, D.~Foss, D.~Wang, D.~Le, D.~Holland, E.~Dowling, E.~Jamil, E.~Montgomery, E.~Presani, E.~Hahn, E.~Wood, E.-T. Le, E.~Brinkman, E.~Arcaute, E.~Dunbar, E.~Smothers, F.~Sun, F.~Kreuk, F.~Tian, F.~Kokkinos, F.~Ozgenel, F.~Caggioni, F.~Kanayet, F.~Seide, G.~M. Florez, G.~Schwarz, G.~Badeer, G.~Swee, G.~Halpern, G.~Herman, G.~Sizov, Guangyi, Zhang, G.~Lakshminarayanan, H.~Inan, H.~Shojanazeri, H.~Zou, H.~Wang, H.~Zha, H.~Habeeb, H.~Rudolph, H.~Suk, H.~Aspegren, H.~Goldman, H.~Zhan, I.~Damlaj, I.~Molybog, I.~Tufanov, I.~Leontiadis, I.-E. Veliche, I.~Gat, J.~Weissman, J.~Geboski, J.~Kohli, J.~Lam, J.~Asher, J.-B. Gaya, J.~Marcus,
  J.~Tang, J.~Chan, J.~Zhen, J.~Reizenstein, J.~Teboul, J.~Zhong, J.~Jin, J.~Yang, J.~Cummings, J.~Carvill, J.~Shepard, J.~McPhie, J.~Torres, J.~Ginsburg, J.~Wang, K.~Wu, K.~H. U, K.~Saxena, K.~Khandelwal, K.~Zand, K.~Matosich, K.~Veeraraghavan, K.~Michelena, K.~Li, K.~Jagadeesh, K.~Huang, K.~Chawla, K.~Huang, L.~Chen, L.~Garg, L.~A, L.~Silva, L.~Bell, L.~Zhang, L.~Guo, L.~Yu, L.~Moshkovich, L.~Wehrstedt, M.~Khabsa, M.~Avalani, M.~Bhatt, M.~Mankus, M.~Hasson, M.~Lennie, M.~Reso, M.~Groshev, M.~Naumov, M.~Lathi, M.~Keneally, M.~Liu, M.~L. Seltzer, M.~Valko, M.~Restrepo, M.~Patel, M.~Vyatskov, M.~Samvelyan, M.~Clark, M.~Macey, M.~Wang, M.~J. Hermoso, M.~Metanat, M.~Rastegari, M.~Bansal, N.~Santhanam, N.~Parks, N.~White, N.~Bawa, N.~Singhal, N.~Egebo, N.~Usunier, N.~Mehta, N.~P. Laptev, N.~Dong, N.~Cheng, O.~Chernoguz, O.~Hart, O.~Salpekar, O.~Kalinli, P.~Kent, P.~Parekh, P.~Saab, P.~Balaji, P.~Rittner, P.~Bontrager, P.~Roux, P.~Dollar, P.~Zvyagina, P.~Ratanchandani, P.~Yuvraj, Q.~Liang, R.~Alao, R.~Rodriguez,
  R.~Ayub, R.~Murthy, R.~Nayani, R.~Mitra, R.~Parthasarathy, R.~Li, R.~Hogan, R.~Battey, R.~Wang, R.~Howes, R.~Rinott, S.~Mehta, S.~Siby, S.~J. Bondu, S.~Datta, S.~Chugh, S.~Hunt, S.~Dhillon, S.~Sidorov, S.~Pan, S.~Mahajan, S.~Verma, S.~Yamamoto, S.~Ramaswamy, S.~Lindsay, S.~Lindsay, S.~Feng, S.~Lin, S.~C. Zha, S.~Patil, S.~Shankar, S.~Zhang, S.~Zhang, S.~Wang, S.~Agarwal, S.~Sajuyigbe, S.~Chintala, S.~Max, S.~Chen, S.~Kehoe, S.~Satterfield, S.~Govindaprasad, S.~Gupta, S.~Deng, S.~Cho, S.~Virk, S.~Subramanian, S.~Choudhury, S.~Goldman, T.~Remez, T.~Glaser, T.~Best, T.~Koehler, T.~Robinson, T.~Li, T.~Zhang, T.~Matthews, T.~Chou, T.~Shaked, V.~Vontimitta, V.~Ajayi, V.~Montanez, V.~Mohan, V.~S. Kumar, V.~Mangla, V.~Ionescu, V.~Poenaru, V.~T. Mihailescu, V.~Ivanov, W.~Li, W.~Wang, W.~Jiang, W.~Bouaziz, W.~Constable, X.~Tang, X.~Wu, X.~Wang, X.~Wu, X.~Gao, Y.~Kleinman, Y.~Chen, Y.~Hu, Y.~Jia, Y.~Qi, Y.~Li, Y.~Zhang, Y.~Zhang, Y.~Adi, Y.~Nam, Yu, Wang, Y.~Zhao, Y.~Hao, Y.~Qian, Y.~Li, Y.~He, Z.~Rait, Z.~DeVito,
  Z.~Rosnbrick, Z.~Wen, Z.~Yang, Z.~Zhao, and Z.~Ma.
\newblock The llama 3 herd of models, 2024.
\newblock URL \url{https://arxiv.org/abs/2407.21783}.

\bibitem[Grigoroglou and Ganea(2022)]{grigoroglou2022language}
M.~Grigoroglou and P.~A. Ganea.
\newblock Language as a mechanism for reasoning about possibilities.
\newblock \emph{Philosophical Transactions of the Royal Society B}, 377\penalty0 (1866):\penalty0 20210334, 2022.

\bibitem[Guo et~al.(2024)Guo, Chen, Wang, Chang, Pei, Chawla, Wiest, and Zhang]{guo2024large}
T.~Guo, X.~Chen, Y.~Wang, R.~Chang, S.~Pei, N.~V. Chawla, O.~Wiest, and X.~Zhang.
\newblock Large language model based multi-agents: A survey of progress and challenges.
\newblock \emph{arXiv preprint arXiv:2402.01680}, 2024.

\bibitem[Hong et~al.(2025)Hong, Troynikov, and Huber]{chromacontentrot}
K.~Hong, A.~Troynikov, and J.~Huber.
\newblock Context rot: How increasing input tokens impacts llm performance, 2025.

\bibitem[Li et~al.(2023)Li, Hammoud, Itani, Khizbullin, and Ghanem]{li2023camel}
G.~Li, H.~Hammoud, H.~Itani, D.~Khizbullin, and B.~Ghanem.
\newblock Camel: Communicative agents for" mind" exploration of large language model society.
\newblock \emph{Advances in Neural Information Processing Systems}, 36:\penalty0 51991--52008, 2023.

\bibitem[Li(2017)]{li2017does}
K.~K. Li.
\newblock How does language affect decision-making in social interactions and decision biases?
\newblock \emph{Journal of Economic Psychology}, 61:\penalty0 15--28, 2017.

\bibitem[Li et~al.(2024)Li, Han, and Ji]{li2024vbloraextremeparameterefficient}
Y.~Li, S.~Han, and S.~Ji.
\newblock Vb-lora: Extreme parameter efficient fine-tuning with vector banks, 2024.
\newblock URL \url{https://arxiv.org/abs/2405.15179}.

\bibitem[Munos et~al.(2024)Munos, Valko, Calandriello, Azar, Rowland, Guo, Tang, Geist, Mesnard, Michi, Selvi, Girgin, Momchev, Bachem, Mankowitz, Precup, and Piot]{munos2024nashlearninghumanfeedback}
R.~Munos, M.~Valko, D.~Calandriello, M.~G. Azar, M.~Rowland, Z.~D. Guo, Y.~Tang, M.~Geist, T.~Mesnard, A.~Michi, M.~Selvi, S.~Girgin, N.~Momchev, O.~Bachem, D.~J. Mankowitz, D.~Precup, and B.~Piot.
\newblock Nash learning from human feedback, 2024.
\newblock URL \url{https://arxiv.org/abs/2312.00886}.

\bibitem[Reimers and Gurevych(2019)]{reimers2019sentencebertsentenceembeddingsusing}
N.~Reimers and I.~Gurevych.
\newblock Sentence-bert: Sentence embeddings using siamese bert-networks, 2019.
\newblock URL \url{https://arxiv.org/abs/1908.10084}.

\bibitem[Shao et~al.(2024)Shao, Wang, Zhu, Xu, Song, Bi, Zhang, Zhang, Li, Wu, and Guo]{shao2024deepseekmathpushinglimitsmathematical}
Z.~Shao, P.~Wang, Q.~Zhu, R.~Xu, J.~Song, X.~Bi, H.~Zhang, M.~Zhang, Y.~K. Li, Y.~Wu, and D.~Guo.
\newblock Deepseekmath: Pushing the limits of mathematical reasoning in open language models, 2024.
\newblock URL \url{https://arxiv.org/abs/2402.03300}.

\bibitem[Shojaee*† et~al.(2025)Shojaee*†, Mirzadeh*, Alizadeh, Horton, Bengio, and Farajtabar]{illusion-of-thinking}
P.~Shojaee*†, I.~Mirzadeh*, K.~Alizadeh, M.~Horton, S.~Bengio, and M.~Farajtabar.
\newblock The illusion of thinking: Understanding the strengths and limitations of reasoning models via the lens of problem complexity, 2025.
\newblock URL \url{https://ml-site.cdn-apple.com/papers/the-illusion-of-thinking.pdf}.

\bibitem[Subramaniam et~al.(2025)Subramaniam, Du, Tenenbaum, Torralba, Li, and Mordatch]{subramaniam2025multiagentfinetuningselfimprovement}
V.~Subramaniam, Y.~Du, J.~B. Tenenbaum, A.~Torralba, S.~Li, and I.~Mordatch.
\newblock Multiagent finetuning: Self improvement with diverse reasoning chains, 2025.
\newblock URL \url{https://arxiv.org/abs/2501.05707}.

\bibitem[Team et~al.(2025)Team, Kamath, Ferret, Pathak, Vieillard, Merhej, Perrin, Matejovicova, Ramé, Rivière, Rouillard, Mesnard, Cideron, bastien Grill, Ramos, Yvinec, Casbon, Pot, Penchev, Liu, Visin, Kenealy, Beyer, Zhai, Tsitsulin, Busa-Fekete, Feng, Sachdeva, Coleman, Gao, Mustafa, Barr, Parisotto, Tian, Eyal, Cherry, Peter, Sinopalnikov, Bhupatiraju, Agarwal, Kazemi, Malkin, Kumar, Vilar, Brusilovsky, Luo, Steiner, Friesen, Sharma, Sharma, Gilady, Goedeckemeyer, Saade, Feng, Kolesnikov, Bendebury, Abdagic, Vadi, György, Pinto, Das, Bapna, Miech, Yang, Paterson, Shenoy, Chakrabarti, Piot, Wu, Shahriari, Petrini, Chen, Lan, Choquette-Choo, Carey, Brick, Deutsch, Eisenbud, Cattle, Cheng, Paparas, Sreepathihalli, Reid, Tran, Zelle, Noland, Huizenga, Kharitonov, Liu, Amirkhanyan, Cameron, Hashemi, Klimczak-Plucińska, Singh, Mehta, Lehri, Hazimeh, Ballantyne, Szpektor, Nardini, Pouget-Abadie, Chan, Stanton, Wieting, Lai, Orbay, Fernandez, Newlan, yeong Ji, Singh, Black, Yu, Hui, Vodrahalli, Greff, Qiu,
  Valentine, Coelho, Ritter, Hoffman, Watson, Chaturvedi, Moynihan, Ma, Babar, Noy, Byrd, Roy, Momchev, Chauhan, Sachdeva, Bunyan, Botarda, Caron, Rubenstein, Culliton, Schmid, Sessa, Xu, Stanczyk, Tafti, Shivanna, Wu, Pan, Rokni, Willoughby, Vallu, Mullins, Jerome, Smoot, Girgin, Iqbal, Reddy, Sheth, Põder, Bhatnagar, Panyam, Eiger, Zhang, Liu, Yacovone, Liechty, Kalra, Evci, Misra, Roseberry, Feinberg, Kolesnikov, Han, Kwon, Chen, Chow, Zhu, Wei, Egyed, Cotruta, Giang, Kirk, Rao, Black, Babar, Lo, Moreira, Martins, Sanseviero, Gonzalez, Gleicher, Warkentin, Mirrokni, Senter, Collins, Barral, Ghahramani, Hadsell, Matias, Sculley, Petrov, Fiedel, Shazeer, Vinyals, Dean, Hassabis, Kavukcuoglu, Farabet, Buchatskaya, Alayrac, Anil, Dmitry, Lepikhin, Borgeaud, Bachem, Joulin, Andreev, Hardin, Dadashi, and Hussenot]{gemmateam2025gemma3technicalreport}
G.~Team, A.~Kamath, J.~Ferret, S.~Pathak, N.~Vieillard, R.~Merhej, S.~Perrin, T.~Matejovicova, A.~Ramé, M.~Rivière, L.~Rouillard, T.~Mesnard, G.~Cideron, J.~bastien Grill, S.~Ramos, E.~Yvinec, M.~Casbon, E.~Pot, I.~Penchev, G.~Liu, F.~Visin, K.~Kenealy, L.~Beyer, X.~Zhai, A.~Tsitsulin, R.~Busa-Fekete, A.~Feng, N.~Sachdeva, B.~Coleman, Y.~Gao, B.~Mustafa, I.~Barr, E.~Parisotto, D.~Tian, M.~Eyal, C.~Cherry, J.-T. Peter, D.~Sinopalnikov, S.~Bhupatiraju, R.~Agarwal, M.~Kazemi, D.~Malkin, R.~Kumar, D.~Vilar, I.~Brusilovsky, J.~Luo, A.~Steiner, A.~Friesen, A.~Sharma, A.~Sharma, A.~M. Gilady, A.~Goedeckemeyer, A.~Saade, A.~Feng, A.~Kolesnikov, A.~Bendebury, A.~Abdagic, A.~Vadi, A.~György, A.~S. Pinto, A.~Das, A.~Bapna, A.~Miech, A.~Yang, A.~Paterson, A.~Shenoy, A.~Chakrabarti, B.~Piot, B.~Wu, B.~Shahriari, B.~Petrini, C.~Chen, C.~L. Lan, C.~A. Choquette-Choo, C.~Carey, C.~Brick, D.~Deutsch, D.~Eisenbud, D.~Cattle, D.~Cheng, D.~Paparas, D.~S. Sreepathihalli, D.~Reid, D.~Tran, D.~Zelle, E.~Noland, E.~Huizenga,
  E.~Kharitonov, F.~Liu, G.~Amirkhanyan, G.~Cameron, H.~Hashemi, H.~Klimczak-Plucińska, H.~Singh, H.~Mehta, H.~T. Lehri, H.~Hazimeh, I.~Ballantyne, I.~Szpektor, I.~Nardini, J.~Pouget-Abadie, J.~Chan, J.~Stanton, J.~Wieting, J.~Lai, J.~Orbay, J.~Fernandez, J.~Newlan, J.~yeong Ji, J.~Singh, K.~Black, K.~Yu, K.~Hui, K.~Vodrahalli, K.~Greff, L.~Qiu, M.~Valentine, M.~Coelho, M.~Ritter, M.~Hoffman, M.~Watson, M.~Chaturvedi, M.~Moynihan, M.~Ma, N.~Babar, N.~Noy, N.~Byrd, N.~Roy, N.~Momchev, N.~Chauhan, N.~Sachdeva, O.~Bunyan, P.~Botarda, P.~Caron, P.~K. Rubenstein, P.~Culliton, P.~Schmid, P.~G. Sessa, P.~Xu, P.~Stanczyk, P.~Tafti, R.~Shivanna, R.~Wu, R.~Pan, R.~Rokni, R.~Willoughby, R.~Vallu, R.~Mullins, S.~Jerome, S.~Smoot, S.~Girgin, S.~Iqbal, S.~Reddy, S.~Sheth, S.~Põder, S.~Bhatnagar, S.~R. Panyam, S.~Eiger, S.~Zhang, T.~Liu, T.~Yacovone, T.~Liechty, U.~Kalra, U.~Evci, V.~Misra, V.~Roseberry, V.~Feinberg, V.~Kolesnikov, W.~Han, W.~Kwon, X.~Chen, Y.~Chow, Y.~Zhu, Z.~Wei, Z.~Egyed, V.~Cotruta, M.~Giang, P.~Kirk,
  A.~Rao, K.~Black, N.~Babar, J.~Lo, E.~Moreira, L.~G. Martins, O.~Sanseviero, L.~Gonzalez, Z.~Gleicher, T.~Warkentin, V.~Mirrokni, E.~Senter, E.~Collins, J.~Barral, Z.~Ghahramani, R.~Hadsell, Y.~Matias, D.~Sculley, S.~Petrov, N.~Fiedel, N.~Shazeer, O.~Vinyals, J.~Dean, D.~Hassabis, K.~Kavukcuoglu, C.~Farabet, E.~Buchatskaya, J.-B. Alayrac, R.~Anil, Dmitry, Lepikhin, S.~Borgeaud, O.~Bachem, A.~Joulin, A.~Andreev, C.~Hardin, R.~Dadashi, and L.~Hussenot.
\newblock Gemma 3 technical report, 2025.
\newblock URL \url{https://arxiv.org/abs/2503.19786}.

\bibitem[Tran et~al.()Tran, Dao, Nguyen, Pham, O’Sullivan, and Nguyen]{tran2501multi}
K.-T. Tran, D.~Dao, M.-D. Nguyen, Q.-V. Pham, B.~O’Sullivan, and H.~D. Nguyen.
\newblock Multi-agent collaboration mechanisms: A survey of llms, 2025.
\newblock \emph{URL https://arxiv. org/abs/2501.06322}.

\bibitem[Webb et~al.(2023)Webb, Holyoak, and Lu]{webb2023emergent}
T.~Webb, K.~J. Holyoak, and H.~Lu.
\newblock Emergent analogical reasoning in large language models.
\newblock \emph{Nature Human Behaviour}, 7\penalty0 (9):\penalty0 1526--1541, 2023.

\bibitem[Wei et~al.(2022)Wei, Tay, Bommasani, Raffel, Zoph, Borgeaud, Yogatama, Bosma, Zhou, Metzler, et~al.]{wei2022emergent}
J.~Wei, Y.~Tay, R.~Bommasani, C.~Raffel, B.~Zoph, S.~Borgeaud, D.~Yogatama, M.~Bosma, D.~Zhou, D.~Metzler, et~al.
\newblock Emergent abilities of large language models.
\newblock \emph{arXiv preprint arXiv:2206.07682}, 2022.

\bibitem[Wu et~al.(2024)Wu, Qiu, Ross, Aky{\"u}rek, Chen, Wang, Kim, Andreas, and Kim]{wu2024reasoning}
Z.~Wu, L.~Qiu, A.~Ross, E.~Aky{\"u}rek, B.~Chen, B.~Wang, N.~Kim, J.~Andreas, and Y.~Kim.
\newblock Reasoning or reciting? exploring the capabilities and limitations of language models through counterfactual tasks.
\newblock Association for Computational Linguistics, 2024.

\bibitem[Yao et~al.(2023)Yao, Zhao, Yu, Du, Shafran, Narasimhan, and Cao]{yao2023reactsynergizingreasoningacting}
S.~Yao, J.~Zhao, D.~Yu, N.~Du, I.~Shafran, K.~Narasimhan, and Y.~Cao.
\newblock React: Synergizing reasoning and acting in language models, 2023.
\newblock URL \url{https://arxiv.org/abs/2210.03629}.

\bibitem[Zhang et~al.(2022)Zhang, Li, Meng, Chang, and Broeck]{zhang2022paradox}
H.~Zhang, L.~H. Li, T.~Meng, K.-W. Chang, and G.~V.~d. Broeck.
\newblock On the paradox of learning to reason from data.
\newblock \emph{arXiv preprint arXiv:2205.11502}, 2022.

\bibitem[Zhu et~al.(2024)Zhu, Dastani, and Wang]{zhu2024survey}
C.~Zhu, M.~Dastani, and S.~Wang.
\newblock A survey of multi-agent deep reinforcement learning with communication.
\newblock \emph{Autonomous Agents and Multi-Agent Systems}, 38\penalty0 (1):\penalty0 4, 2024.

\end{thebibliography}

\newpage
\appendix

\end{document}